\documentclass[lettersize,journal]{IEEEtran}

\usepackage{amsmath,amsfonts}
\usepackage{array}
\usepackage{textcomp}
\usepackage{stfloats}
\usepackage{url}
\usepackage{verbatim}
\usepackage{graphicx}
\usepackage{hyperref}
\usepackage{amssymb}
\usepackage{booktabs}
\usepackage{algorithm}
\usepackage{multirow}
\usepackage{makecell}
\usepackage{subfigure}
\usepackage{color}
\usepackage{algorithm}
\usepackage{algpseudocode}

\usepackage{caption}
\usepackage{xcolor}

\usepackage[table]{xcolor}

\newcommand{\revI}[1]{#1}
\newcommand{\revII}[1]{#1}
\newcommand{\revIII}[1]{#1}

\usepackage{forloop}
\newcounter{ct}

\def\BibTeX{{\rm B\kern-.05em{\sc i\kern-.025em b}\kern-.08em
    T\kern-.1667em\lower.7ex\hbox{E}\kern-.125emX}}
\usepackage{balance}

\usepackage{color}
\usepackage{bm}
\usepackage{hyperref}
\usepackage{amsmath,amsfonts}
\usepackage{blindtext}
\usepackage{listings}
\usepackage{mathtools}

\DeclarePairedDelimiterX{\infdivx}[2]{(}{)}{%
	#1\;\delimsize\|\;#2%
}

\newcommand{\vect}[1]{\bm{#1}}

\newcommand{\x}{\xv}
\newcommand{\D}{\Dv}

\newcommand{\dm}{\mathrm{d}}

\newcommand{\N}{\mathcal{N}}

\newcommand{\dlambda}{\mathrm{d}\lambda}
\newcommand{\dxv}{\mathrm{d}\xv}

\newcommand{\dt}{\mathrm{d}t}

\newcommand{\epsilonv}{\vect\epsilon}

\newcommand{\xv}{\vect x}

\newcommand{\Dv}{\vect D}

\newcommand{\Iv}{\vect I}

\makeatletter
\newcommand{\newreptheorem}[2]{%
\newenvironment{rep#1}[1]{%
 \def\rep@title{#2 \ref{##1}}%
 \begin{rep@theorem}}%
 {\end{rep@theorem}}}
\makeatother

\newreptheorem{proposition}{Proposition}
\newreptheorem{theorem}{Theorem}
\newreptheorem{lemma}{Lemma}

\begin{document}

\title{SynBoost: A Synergistic Framework for Fast Sampling of Diffusion Models}

\author{Hu Yu, Hao Luo, Xueyang Fu, Jie Huang, Fan Wang, Feng Zhao, \textit{IEEE, Member}
\thanks{
This work was supported by the Anhui Provincial Natural Science Foundation under Grant 2108085UD12. We acknowledge the support of GPU cluster built by MCC Lab of Information Science and Technology Institution, USTC. The AI-driven experiments, simulations and model training were performed on the robotic AI-Scientist platform of Chinese Academy of Sciences. (Corresponding author: Feng Zhao.)

H Yu, X Fu, J Huang, and F Zhao are with MoE Key Laboratory of Brain-inspired Intelligent Perception and Cognition, University of Science and Technology of China. H Luo and F Wang are with Alibaba DAMO Academy for Discovery, Adventure, Momentum and Outlook. 

This paper has supplementary downloadable material available at http://ieeexplore.ieee.org., provided by the author. The material includes more explanations, experiments and analyses. This material is 3.4 MB in size.

}
}

\markboth{Journal of \LaTeX\ Class Files,~Vol.~18, No.~9, September~2020}%
{How to Use the IEEEtran \LaTeX \ Templates}

\maketitle

\begin{abstract}
    Diffusion probabilistic models (DPMs) have demonstrated remarkable success in visual generation. However, their iterative sampling mechanism results in slow inference speeds. While reducing sampling steps offers an intuitive acceleration strategy, it introduces significant discretization error. Existing fast samplers have made substantial progress in mitigating this error through high-order solvers, yet further optimization appears constrained. This limitation prompts a critical question: can sampling efficiency be advanced beyond current paradigms?
    In this work, we re-examine the composition of sampling errors and identify two distinct components: the well-studied discretization error and the under-explored approximation error. By implementing a dual-error disentanglement strategy, we elucidate the dynamic interplay between these error types across sampling steps. Building on this empirical analysis, we propose SynBoost, a unified and training-free acceleration framework that simultaneously addresses both error sources to minimize total sampling error. \revII{Concretely, it mitigates approximation error by blending the current noise estimation with a more accurate prediction from a larger timestep.}
    SynBoost seamlessly integrates with existing samplers, substantially enhancing their speed and output quality, particularly in regimes with extremely few steps. We validate our framework through extensive experiments across unconditional and conditional generation tasks, encompassing both pixel-space and latent-space DPMs.
    
\end{abstract}

\begin{IEEEkeywords}
Diffusion model, inference acceleration, error decomposition.
\end{IEEEkeywords}

\section{Introduction}
\label{sec:intro}

Diffusion probabilistic models (DPMs) \cite{sohl2015deep,ho2020denoising,song2020score} have demonstrated impressive success across a broad spectrum of tasks, including image synthesis \cite{dhariwal2021diffusion,rombach2022high,ramesh2022hierarchical,saharia2022photorealistic}, image editing \cite{meng2021sdedit}, video generation \cite{ho2022video,blattmann2023stable}, voice synthesis \cite{chen2020wavegrad}, etc.
Compared with other generative models such as GANs \cite{goodfellow2014generative} and VAEs \cite{kingma2013auto}, DPMs not only exhibit superior sample quality but also benefit from a more robust training methodology and an advanced guided sampling technique. However, the inference of DPMs usually requires multiple model evaluations (NFEs), which hinders their practical deployment.

\begin{figure*}[t]
    \begin{center}
	\includegraphics[width=0.98\linewidth]{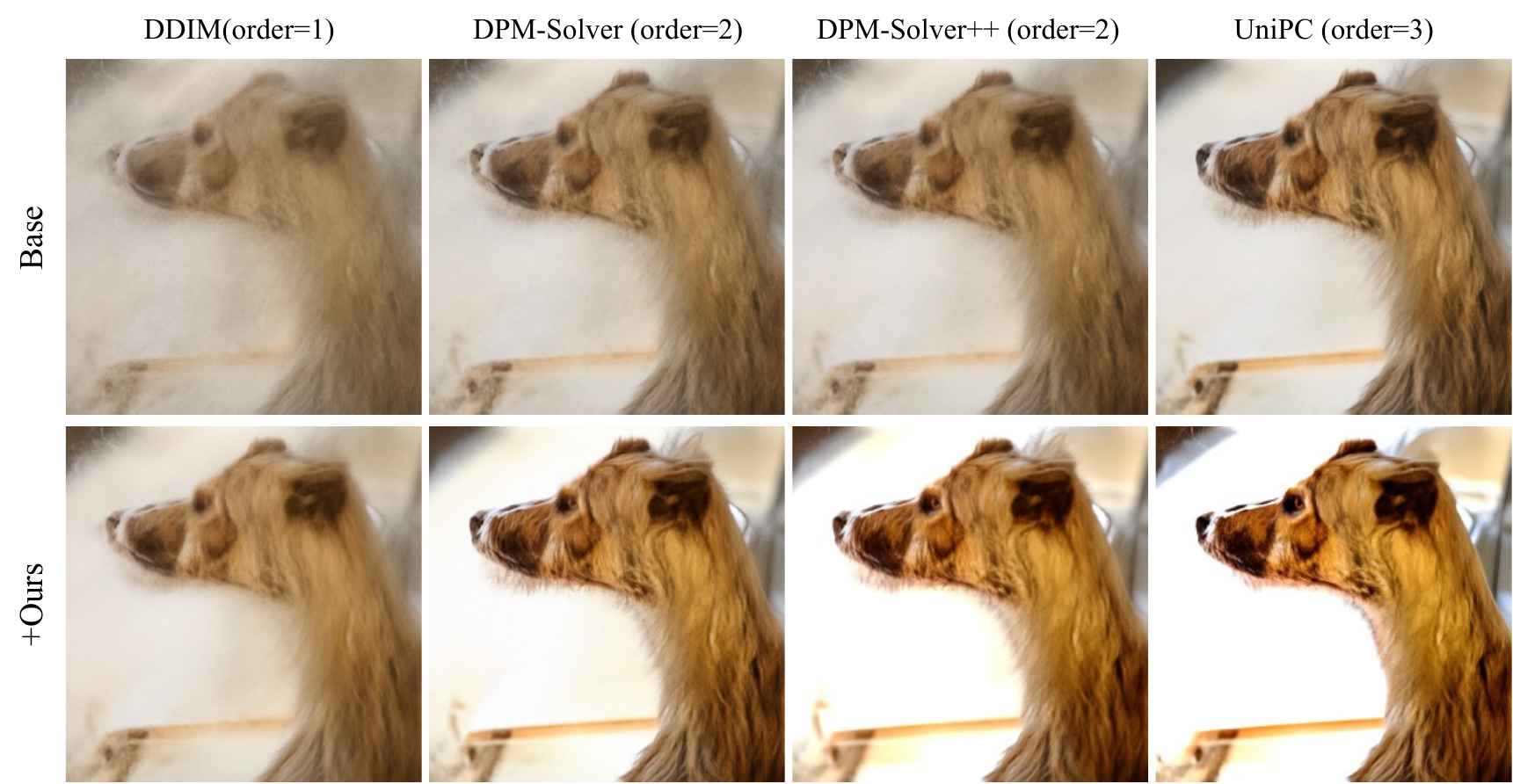}
    \end{center}
    \setlength{\abovecaptionskip}{-0.1cm}
    \setlength{\belowcaptionskip}{-0.3cm}
    \caption{Qualitative comparisons between existing solvers and our method with class condition. All images are generated by sampling from ADM \cite{dhariwal2021diffusion} trained on ImageNet 256×256 with only 7 number of function evaluations (NFEs). We show that our proposed method can significantly elevate the sample quality with better details and contrast than previous base samplers.} 
    \label{fig:demo_pixel}
\end{figure*}

Recently, there has been a surge in endeavors to expedite the sampling processes of DPMs \cite{salimans2022progressive,meng2023distillation,yu2025comp,qin2025truncate,song2023consistency,song2020denoising,lu2022dpm,zhang2022fast,yu2023debias}, which are broadly categorized into training-based model distillation and training-free fast sampling approaches. Distillation-based techniques, notable for facilitating generation in a minimal number of steps, are somewhat curtailed by a complex distillation procedure and the necessity for model-specific distillation. For example, Comp-Diff \cite{yu2025comp} proposes the pruning and distillation framework for compressing diffusion models.
Conversely, fast samplers \cite{song2020denoising,zhang2022gddim,liu2022pseudo,lu2022dpm,zhang2022fast,lu2022dpm++,zhao2023unipc,wang2023boosting,zhang2023lookahead,xu2023restart,zheng2023dpm} enjoy wider popularity for their inherent training-free quality, allowing seamless integration with readily available pre-trained DPMs. 
These methods predominantly leverage probability flow ordinary differential equations (ODEs) and can be formulated in a general continuous exponential integrator form.
Their main strategy centers on minimizing discretization errors that emerge from approximating such continuous integration with large step size discretization (small sampling steps), and potentially reaching a plateau in terms of optimization.
However, an intriguing question arises: \textit{besides minimizing the discretizaton error, is it possible to further elevate the sampling speed and quality with fewer-step inference?}

In this paper, we start with a detailed examination of the constituents of the total sampling error, discerning it into discretization and approximation errors. The former part originates from the discretization of the continuous exponential integrator, whereas the latter part emerges due to the neural network's imprecise estimation of the ground truth vector field (score function).  
Prevailing fast solvers predominantly address the discretization error, inadvertently ignoring the approximation error, thus leaving space for enhanced inference acceleration.
Further, we empirically introduce a dual-error disentanglement strategy aimed at isolating these two sub-errors from the aggregate sampling error, subsequently elucidating their interrelations and their distinct associations with the step. 
Our findings indicate that both errors are of comparable magnitude, underscoring the criticality and significance of reducing the approximation error as part of our contribution. Moreover, we experimentally discover that the approximation error monotonically decreases as step $t$ increases, a relationship that is pivotal for guiding the reduction of approximation error in later designs.

With these insights, we propose an unified and training-free acceleration framework, SynBoost \footnote{\revI{The name ``SynBoost'' in our work stands for ``\textbf{Syn}ergistic \textbf{Boost}ing''.}}, for the fast sampling of DPMs by taking both discretization and approximation errors into consideration to further reduce the total sampling error. To mitigate the discretization error, SynBoost readily incorporates existing ODE solver techniques. 
\revII{To address the approximation error, we design a plug-and-play reduction strategy that integrates fluidly with current ODE solvers, thereby facilitating further acceleration. Specifically, we propose a training-free mixing strategy that partially substitutes the noise estimation at the current timestep with a highly accurate prediction from a preceding, larger timestep.
Furthermore, we elucidate the process of integrating this approximation error reduction strategy into several representative fast solvers, including DDIM \cite{song2020denoising}, DPM-Solver \cite{lu2022dpm}, and DPM-Solver++ \cite{lu2022dpm++}, spanning various orders and prediction modes. }
\revII{Note} that this strategy is analyzed comprehensively and empirically, rather than pursuing strict theoretical rigor.

The efficacy of our SynBoost framework is rigorously substantiated through comprehensive experiments that span a diverse range of models, datasets, solvers, and sampling steps. Specifically, these experiments are structured to encompass two sampling types (unconditional and conditional generation), two condition types (class and text conditions), two sampling spaces (pixel-space and latent-space DPMs), samplers of different orders (1-order DDIM, 2-order DPM-Solver and DPM-Solver++), different prediction modes (noise and data predictions), and various sampling steps. 
SynBoost significantly improves the sampling quality and efficiency over previous solvers on all the conducted experiments, especially with extremely limited sampling steps. 
Qualitative comparisons, as depicted in Figure \ref{fig:demo_pixel}, reveal a consistent advantage of our method in producing images of superior structure, details, and color contrast over the base solvers.

\section{Related Work}
\subsection{Diffusion Probabilistic Models}
Diffusion probabilistic models (DPMs) \cite{sohl2015deep,ho2020denoising,song2020score} transform complex data distributions into simple noise distributions and learn to recover data from noise. The forward diffusion process starts from a clean data sample $\x_0$ and repeatedly injects Gaussian noise to a simple normal distribution $ \x_T\sim \N(\vect{0},\sigma^2\Iv)$ at time $T > 0$ for some $\sigma > 0$. The corresponding transition kernel $q_{t \mid 0}(\x_{t} \mid \x_{0})$ is as follows: 
\begin{equation}
	\label{eq:1}
	q_{t \mid 0}(\x_{t} \mid \x_{0})= \N(\x_t | \alpha_t\x_0, \sigma_t^2\Iv),  \\
\end{equation}
where the \textit{signal-to-noise-ratio} (SNR) equals $\alpha_{t}^{2} / \sigma_{t}^{2}$.

\noindent \textbf{Training process.}
Diffusion models are trained by optimizing a variational lower bound (VLB). For each step $t$, the denoising score matching loss is the distance between two Gaussian distributions, written as:
\begin{equation}
	\label{eq:2}
	\min _{\theta} \mathbb{E}_{\x_{0}, \epsilonv, t} \Big[\omega(t)\|\epsilonv_\theta(\x_t,t)-\epsilonv\|_2^2\Big], \\
\end{equation}
where $\omega(t)$ is weighting function, $\epsilonv\sim q(\epsilonv)=\N(\epsilonv|\vect{0},\Iv)$, and $\x_t=\alpha_t \x_0+\sigma_t \epsilonv$.

\noindent \textbf{ODE-based sampling process.}
Sampling from DPMs can be achieved by solving the following diffusion ODE \cite{song2020score}:
\begin{equation}
	\label{eq:3}
	\frac{\dxv_{t}}{\dt}=f(t) \x_{t}+\frac{g^{2}(t)}{2 \sigma_{t}} \epsilonv_{\theta}(\x_{t}, t), \quad \x_T\sim \N(\vect{0},\sigma^2\Iv) , \\
\end{equation}
where the coefficients $f(t)=\frac{\dm \log \alpha_{t}}{\dt}$, $g^{2}(t)=\frac{\dm \sigma_{t}^{2}}{\dt}-2 \frac{\dm \log \alpha_{t}}{\dt} \sigma_{t}^{2}$.

\subsection{Fast Sampling of DPMs}
A large step size in stochastic differential equations (SDEs) violates the randomness of the Wiener process \cite{Kloeden1992numerical} and often causes non-convergence. 
Certain methods \cite{guo2024gaussian,gonzalez2024seeds} propose to accelerate SDE solvers but still lag behind ODE solvers in speed. \revII{AdaSDE \cite{wangadaptive} achieves strong results in the low-NFE regime by unifying ODE efficiency and SDE error resilience through a learnable error-correction coefficient.}
Restart sampling \cite{xu2023restart} combines the SDE and ODE to boost sampling quality, and also mentions the concept of the approximation error, but it neglects to present the reason, disentanglement, and impact of this error. 
\revII{EPD-Solver \cite{zhu2025distilling} leverages parallel gradient evaluations and lightweight distillation to effectively mitigate truncation errors.}
Chen et al. \cite{chen2023score} explored to decompose the score function of the linear subspace data, but under the low-dimensional linear subspace assumption.
Hunter et al. \cite{hunter2023fast} reduced the cost of calling the U-net via reusing the attention, while keeping the total steps unchanged.
For faster sampling, the probability flow ODE \cite{song2020score} is usually considered as a better choice.
Recent works \cite{lu2022dpm,zhang2022fast,lu2022dpm++} find that ODE solvers built on exponential integrators \cite{hochbruck2010exponential} appear to have faster convergence than directly solving the diffusion ODE Equation \ref{eq:3}. The solution $\x_{t_i}$ of the diffusion ODE given the initial value $\x_{t_{i-1}}$ can be computed as \cite{lu2022dpm}:
\begin{equation}
	\label{eq:4}
	\x_{t_i}=\frac{\alpha_{t_i}}{\alpha_{t_{i-1}}} \x_{t_{i-1}} - \alpha_{t_i} \int_{\lambda_{t_{i-1}}}^{\lambda_{t_i}} e^{-\lambda} \hat{\epsilonv}_{\theta}(\hat{\x}_{\lambda}, \lambda) \dlambda , \\
\end{equation}
where the ODE is changed from the time (t) domain to the log-SNR ($\lambda$) domain by the change-of-variables formula, and the notation $\hat{\epsilonv}_{\theta}$ and $\hat{\x}_{\lambda}$ denote change-of-variables. Based on the exponential integrator, existing ODE solvers approximate $\hat{\epsilonv}_{\theta}(\hat{\x}_{\lambda}, \lambda)$ via Taylor expansion at time $\lambda_{t_{i-1}}$. 
\begin{equation}
    \label{eq:5}
    \begin{aligned}
    \x_{t_{i}}&=\frac{\alpha_{t_{i}}}{\alpha_{t_{i-1}}} \x_{t_{i-1}}-\alpha_{t_{i}} \sum_{n=0}^{k-1} \hat{\epsilonv}_{\theta}^{(n)}(\hat{\x}_{\lambda_{t_{i-1}}}, \lambda_{t_{i-1}}) \int_{\lambda_{t_{i-1}}}^{\lambda_{t_{i}}} \\
    & e^{-\lambda} \frac{(\lambda-\lambda_{t_{i-1}})^{n}}{n !} \dlambda + \mathcal{O}(h_{i}^{k+1}), \\
    \end{aligned}
\end{equation}
where $k$ denotes the order of the Taylor expansion, also known as the order of the solver, and $h_{i} = \lambda_{t_{i}} - \lambda_{t_{i-1}}$ is the step size. Obviously, high order solver reduces the discretization error to $\mathcal{O}(h_{i}^{k+1})$.
Further, it can be simplified into a general form as follows:
\begin{equation}
	\label{eq:6}
	\x_{t_{i}}=\frac{\alpha_{t_{i}}}{\alpha_{t_{i-1}}} \x_{t_{i-1}}-\sigma_{t_{i}}(e^{h_i}-1) \D_{t_{i}}. \\
\end{equation}
Existing ODE solvers only differ in $\D_{t_{i}}$. 
For the first-order DDIM solver, $\D_{t_{i}} = \epsilonv_{\theta}(\hat{\x}_{t_{i-1}}, t_{i-1})$.
EDM \cite{karras2022elucidating} finds that the discretization error is smaller at larger noise level.
DPM-Solver++ \cite{lu2022dpm++} considers rewriting Equation \ref{eq:4} using $\hat{\x}_{\theta}$ instead of $\hat{\epsilonv}_{\theta}$. 
UniPC \cite{zhao2023unipc} proposes a predictor-corrector method to refine the prediction. 
A common thread in these approaches is that they attempt to reduce discretization error part via high order Taylor approximation in Equation \ref{eq:5}, while ignoring the approximation error induced in Equation \ref{eq:4} when replacing score function with network approximation $\hat{\epsilonv}_{\theta}(\hat{\x}_{\lambda}, \lambda)$.
\revII{It is worth noting that our framework is also theoretically orthogonal and compatible with linear multi-step methods like iPNDM \cite{liu2022pseudo}, though exponential integrators remain our primary focus due to their dominance in the extreme low-step regime.}

\section{Method}
\subsection{Motivation and Insights}
We delve into the evolution from the exact solution of diffusion ordinary differential equations (ODEs) to its practical implementation form as illustrated in Equation \ref{eq:6}:

\begin{equation}
    \label{eq:7}
    \begin{aligned}
    x_{t_i} &= \frac{\alpha_{t_i}}{\alpha_{t_{i-1}}}x_{t_{i-1}} - \alpha_{t_i}\int_{\lambda_{t_{i-1}}}^{\lambda_{t_i}}e^{-\lambda}[-\sigma_t\nabla_x\log q_t(x_t)]d\lambda, \\
    &= \frac{\alpha_{t_i}}{\alpha_{t_{i-1}}}x_{t_{i-1}} - \alpha_{t_i}\int_{\lambda_{t_{i-1}}}^{\lambda_{t_i}}e^{-\lambda}\hat{\epsilon}_\theta(\hat{x}_\lambda,\lambda)d\lambda + \mathcal{E}_{app},\\
    &= \frac{\alpha_{t_i}}{\alpha_{t_{i-1}}}x_{t_{i-1}} - \sigma_{t_i}(e^{h_i}-1)D_{t_i} + \mathcal{E}_{app} + \mathcal{E}_{dis}. \\
      \end{aligned}
\end{equation}

Here, $\D_{t_{i}}$ signifies the polynomial of $\hat{\epsilonv}_{\theta}(\hat{\x}_{\lambda}, \lambda)$, mirroring varying degrees of discrete approximations to the continuous integral. \revII{The above formulation reveals that the induced error consists of two parts: discretization error ($\mathcal{E}_{dis}$) and approximation error ($\mathcal{E}_{app}$).}
The approximation error arises from the substitution of the precise score function $-\sigma_{t} \nabla_{\x} \log q_{t}(\x_{t})$ with the network's approximation $\hat{\epsilonv}_{\theta}(\hat{\x}_{\lambda}, \lambda)$ due to the denoising network's imprecise score function estimation.
Subsequently, discretization error emerges when simulating the continuous integral through discrete implementation. While previous fast samplers predominantly aim at diminishing discretization error via higher-order Taylor expansions, introducing variously ordered samplers such as 1-order DDIM and 2-order DPM-Solver, they overlook the criticality of approximation error.
It is noticeable that while some prior methods \cite{xu2023restart,hunter2023fast} may recognize the presence of approximation error, they conclude the exploration at this point. In stark contrast, this derivation is the start of our study. We conduct comprehensive analyses and studies to thoroughly unlock the approximation error, and proceed to integrating this error into existing solvers for further acceleration with an unified and general ODE-based acceleration framework.

\begin{figure*}[t]
    \begin{center}
	\includegraphics[width=0.98\linewidth]{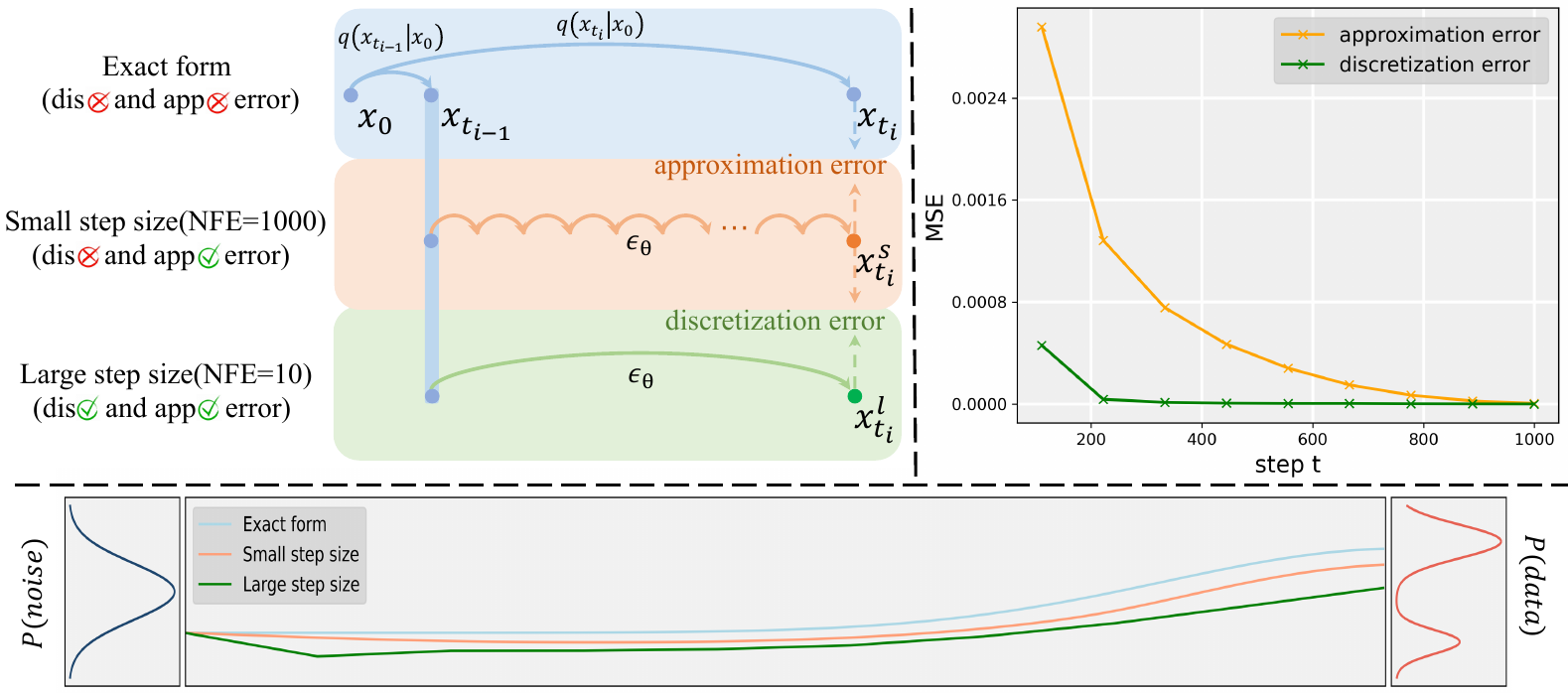}
    \end{center}
    \setlength{\abovecaptionskip}{-0.2cm}
    \setlength{\belowcaptionskip}{-0.4cm}
    \caption{(Left): Dual-Error Disentanglement. Empirically, for the identical time period, we map out three distinctive distribution transition processes, each subjected to a unique combination of errors.
    The top one represents exact data distribution and is free of errors. The middle one is liberated from discretization error owing to an extremely fine-grained step size (NFE=1000). The bottom one, with a coarse step size (NFE=10), succumbs to both errors. (Right): The MSE curve of approximation and discretization errors along the step $t$, \revIII{evaluated on the unconditional generation task using the ImageNet dataset. Note that this monotonically decreasing trend remains highly consistent across different generation settings (e.g., classifier-guided conditional generation, as detailed in Figure \ref{fig:rebuttal_error} of the experimental part).} (Bottom): Overview of ODE-based generation process mapping noise to data distribution, where smaller errors correspond to higher probability density region.}
    \label{fig:error}
\end{figure*}

\subsection{Dual-Error Disentanglement}
With the above insights on the total sampling error, we then delve into dissecting the properties of these two errors. To achieve this, we introduce a dual-error disentanglement strategy, effectively isolating these sub-errors as depicted in Figure \ref{fig:error}. 
Specifically, within the time interval $[t_{i-1},t_i]$, we construct three distinct transition processes, each subjected to varying levels of sampling error. Initially, we define the exact data distributions $\mathcal{P}(\x_{t_{i-1}})$ and $\mathcal{P}(\x_{t_{i}})$, which are free of both approximation and discretization errors. These distributions are derived from the pristine image distribution $\mathcal{P}(\x_0)$, employing the transition kernel outlined in Equation \ref{eq:1}. This procedure establishes an optimal, error-free transition from distribution $\mathcal{P}(\x_{t_{i-1}})$ to $\mathcal{P}(\x_{t_{i}})$. 
We then formulate a second distribution transition from $\mathcal{P}(\x_{t_{i-1}})$ to $\mathcal{P}(\x_{t_{i}}^s)$, which is largely afflicted by approximation error, alleviating the effect of discretization error. To this end, we adopt an extremely small step size (NFE=1000) to minimize the discretization error when approximating the continuous integral. Subsequently, we chart the third distribution transition from $\mathcal{P}(\x_{t_{i-1}})$ to $\mathcal{P}(\x_{t_{i}}^l)$, suffering from both discretization and approximation errors, facilitated by a large step size (i.e., NFE=10).

Originating from the identical data distribution $\mathcal{P}(\x_{t_{i-1}})$, we now derive three distributions $\mathcal{P}(\x_{t_{i}})$, $\mathcal{P}(\x_{t_{i}}^s)$, and $\mathcal{P}(\x_{t_{i}}^l)$, thereby disentangling approximation and discretization errors within the time range $[t_{i-1},t_i]$. Notably, the discrepancy between $\mathcal{P}(\x_{t_{i}})$ and $\mathcal{P}(\x_{t_{i}}^s)$ illuminates the approximation error, whereas the divergence between $\mathcal{P}(\x_{t_{i}}^s)$ and $\mathcal{P}(\x_{t_{i}}^l)$ manifests the discretization error. We adopt the following formulation to approximately represent these two errors. The rationality of adopting MSE to measure distribution distance is explained in the Analyses part of the Experiment section. 
\begin{equation}
    \label{eq:error}
    \begin{aligned}
    E_{app}&=MSE(\x_{t_{i}}^s, \x_{t_{i}}), \\
    E_{dis}&=MSE(\x_{t_{i}}^s, \x_{t_{i}}^l). \\
    \end{aligned}
\end{equation}

An illustrative MSE curve of these errors across the step $t$ is depicted in Figure \ref{fig:error}. Specifically, we divide the total generation period to 9 segments, with each covering the period size of 111. Therefore, NFE=111 and NFE=1 respectively represent the small step manner and large step manner. The right part of Figure \ref{fig:error} is then plotted with these 9 segments.
Analysis of the MSE curve yields three pivotal insights: (1) Both errors are of comparable magnitude and the approximation error surpasses the discretization error at most timesteps, highlighting its significant influence in the sampling process and underscoring the criticality of reducing the approximation error for accelerating sampling. (2) The discretization error declines as the step $t$ increases, which is also consistent with the conclusion in EDM \cite{karras2022elucidating}.
(3) The approximation error exhibits a strict decline as the step $t$ increases, a principle that subsequently instructs the design of our approximation error reduction strategy.

\subsection{Acceleration Framework-SynBoost}
In this section, we introduce our unified and training-free SynBoost framework, crafted to synergize with prevailing fast samplers while concurrently mitigating the approximation error for augmented acceleration. For the discretization error, we seamlessly incorporate the Taylor approximation mechanism employed by preceding fast ODE solvers. 
For the approximation error, we formulate a reduction strategy by leveraging the property of the approximation error that it monotonically decreases with advancing step $t$. This characteristic, where the network's estimation of the actual score function gains accuracy at larger step, guides us to partially substitute the noise prediction at the current step $t_i$ with that of a preceding, larger step $\tau$, given its diminished approximation error.
\begin{equation}
	\label{eq:8}
	\epsilonv_{\theta}^{new}(\x_{t_i}, t_i) = (1+c)\epsilonv_{\theta}(\x_{t_i}, t_i) - c \epsilonv_{\theta}(\x_{\tau}, \tau), \\
\end{equation}
where $c$ is the mixing coefficient, and step $\tau$ is larger than current step $t_i$. 
\revI{We provide comprehensive theoretical and empirical justifications for this formulation in the supplementary material. Specifically, we detail: 1) its strict mathematical derivation from the exact DDIM sampling step; 2) the theoretical expansion of the direct substitution baseline; and 3) empirical evidence demonstrating the severe numerical instability (e.g., generation failure) inherently caused by direct substitution.}
The mixing coefficient $c$ should monotonically increase as current step $t_i$ decreases, echoing the tendency that smaller steps correspond to larger approximation error and thus necessitates more substantial rectification. Besides, we adopt the initial noise $\x_T$ as $\epsilonv_{\theta}(\x_{\tau}, \tau)$. \revI{Note that this does not introduce extra NFEs, since the prediction $\epsilon_{\theta}(x_T, T)$ is computed once at the first step and cached in memory.} A detailed examination and ablation study pertaining to the configurations of $c$ and $\tau$ are presented in Sec. \ref{subsec:analysis}.
Besides, while Equation \ref{eq:8} is expressed within the context of noise prediction, it is equally applicable to data prediction due to the equivalent transformation between these modes. Note that, SynBoost only needs one NFE per step. 

The rationality of Equation \ref{eq:8} can also be verified 
by its similarity to the multi-step method \cite{liu2022pseudo}, which averages the history noise prediction to reduce error.

SynBoost can be seamlessly integrated into existing fast ODE solvers to achieve further speedup. We provide a thorough mathematical integration process of this approximation error reduction strategy into existing solvers. For instance, we select three common and representative fast ODE samplers: DDIM, DPM-Solver, and DPM-Solver++, spanning various orders and prediction modes. \revI{Furthermore, our framework is fundamentally compatible with other state-of-the-art exponential integrators such as UniPC \cite{zhao2023unipc}. Since SynBoost operates by refining approximation error, it is completely orthogonal to UniPC's multi-step discretization rules, allowing for seamless plug-and-play integration without structural conflicts. We also present the integration of our method with UniPC in the experimental part.}
Here, we again emphasize that Equation \ref{eq:6} is the general formulation of the ODE-based inference, with the only difference between different samplers in $\D_{t_{i}}$.

\noindent \textbf{Ours-DDIM.}
DDIM is the classical ODE sampler that proposes a non-Markovian diffusion process following the same training objective as DDPMs. This non-Markovian process leads to deterministic generation and faster speed.
The sampling formulation of DDIM is as follows:
\begin{equation}
	\label{eq:9}
	\x_{t_i}^{base}= \alpha_{t_{i}} \x_{\theta}(\x_{t_{i-1}}, t_{i-1}) + \sigma_{t_{i}} \epsilonv_{\theta}(\x_{t_{i-1}}, t_{i-1}). \\
\end{equation}
Further, we can rewrite the above equation in the unified form of Equation \ref{eq:6} with following $\D_{t_{i}}$:
\begin{equation}
	\label{eq:10}
	\D_{t_{i}}^{base} = \epsilonv_{\theta}(\x_{t_{i-1}}, t_{i-1}). \\
\end{equation}
To reduce the approximation error, we replace the noise estimation part in Equation \ref{eq:10} with Equation \ref{eq:8}.

\begin{equation}
	\label{eq:12}
	\D_{t_{i}}^{ours} = (1 + c) \epsilonv_{\theta}(\x_{t_{i-1}}, t_{i-1}) - c \epsilonv_{\theta}(\x_{\tau, \tau}).\\
\end{equation}

\noindent \textbf{Ours-DPM-Solver.}
We apply multi-step, thresholding strategy \cite{lu2022dpm++} and second-order to DPM-Solver, and get the base version termed as DPM-Solver(2M). Note that the only difference between DPM-Solver(2M) and DPM-Solver++(2M) is the prediction mode.

DPM-Solver reveals that diffusion ODEs have a semi-linear structure and derives the formulation of the solutions by analytically computing the linear part of the solutions, avoiding the corresponding discretization error.
Concretely, DPM-Solver(2M) can be directly written in the formation of Equation \ref{eq:6} with the corresponding $\D_{t_{i}}$:
\begin{equation}
	\label{eq:13}
	\D_{t_{i}}^{base} = \epsilonv_\theta(\x_{t_{i-1}}, t_{i-1}) + a_1 [ \epsilonv_\theta(\x_{t_{i-1}}, t_{i-1}) - \epsilonv_\theta(\x_{t_{i-2}}, t_{i-2}) ], \\
\end{equation}
where $a_1$ is the coefficient for the second part of DPM-Solver. Then, we adopt a similar way as in previous DDIM part to reduce the approximation error.
\begin{equation}
    \label{eq:14}
    \begin{aligned}
    \D_{t_{i}}^{ours} &= \left[ (1+c)\epsilonv_\theta(\x_{t_{i-1}}, t_{i-1}) - c \epsilonv_\theta(\x_{\tau}, \tau)\right]+ \\
    &a_1 \left[ \epsilonv_\theta(\x_{t_{i-1}}, t_{i-1}) - \epsilonv_\theta(\x_{t_{i-2}}, t_{i-2}) \right]. \\
    \end{aligned}
\end{equation}

\noindent \textbf{Ours-DPM-Solver++.}
DPM-Solver++ is the SOTA and default sampler in the stable diffusion model. It finds that previous high-order fast samplers suffer from instability issue, and solves the diffusion ODE with the data prediction model. Due to employing different prediction modes, DPM-Solver++ reformulates Equation \ref{eq:6} as follows:
\begin{equation}
	\label{eq:15}
        \x_{t_{i}} = \frac{\sigma_{t_{i}}}{\sigma_{t_{i-1}}} \x_{t_{i-1}}-\alpha_{t_{i}}(e^{-h_{i}}-1) \D_{t_i},  \\
\end{equation}
where $\D_{t_i}$ is expressed in the data-prediction mode $\x_{\theta}(\x_{t_{i-1}}, t_{i-1})$ as follows:
\begin{equation}
    \label{eq:16}
    \begin{aligned}
    \D_{t_{i}}^{base} &= \x_{\theta}(\x_{t_{i-1}}, t_{i-1}) + a_2 [ \x_{\theta}(\x_{t_{i-1}}, t_{i-1}) \\ 
    &- \x_{\theta}(\x_{t_{i-2}}, t_{i-2}) ], \\
    \end{aligned}
\end{equation}
where $a_2$ is the coefficient for the second part of DPM-Solver++. The data prediction $\x_{\theta}(\x_{t_{i-1}}, t_{i-1})$ and noise prediction $\epsilonv_{\theta}(\x_{t_{i-1}}, t_{i-1})$ can be mutually derived from each other with Equation \ref{eq:1}.
\begin{equation}
	\label{eq:17}
	\x_{t_{i-1}} = \alpha_{t_{i-1}} \x_{\theta}(\x_{t_{i-1}}, t_{i-1}) + \sigma_{t_{i-1}} \epsilonv_{\theta}(\x_{t_{i-1}}, t_{i-1}). \\
\end{equation}
Therefore, we first convert the first order part $\x_{\theta}(\x_{t_{i-1}},t_{i-1})$ in Equation \ref{eq:16} to noise prediction $\epsilonv_{\theta}(\x_{t_{i-1}}, t_{i-1})$, and apply Equation \ref{eq:8} to the converted $\epsilonv_{\theta}(\x_{t_{i-1}}, t_{i-1})$, and finally convert it back to data prediction.

\section{Experiments}
\label{sec:exp}
In this section, we show that our method can significantly boost the sampling quality and speed of existing solvers through extensive experiments. We employ FID and the human preference model HPD v2 \cite{wu2023human} for comprehensive evaluation. 
\revII{Following the setting of recent training-free samplers \cite{lu2022dpm++, zhao2023unipc}, all values are reported using 10k images unless specifically mentioned. This ensures computational efficiency across our extensive ablation sweeps while maintaining strictly fair across all baseline methods.}
We first present main results in Section \ref{subsec:main} and provide more analyses in Section \ref{subsec:analysis}.

\begin{figure*}[p]

    \begin{minipage}[b]{0.98\textwidth}
        \includegraphics[width=0.98\linewidth]{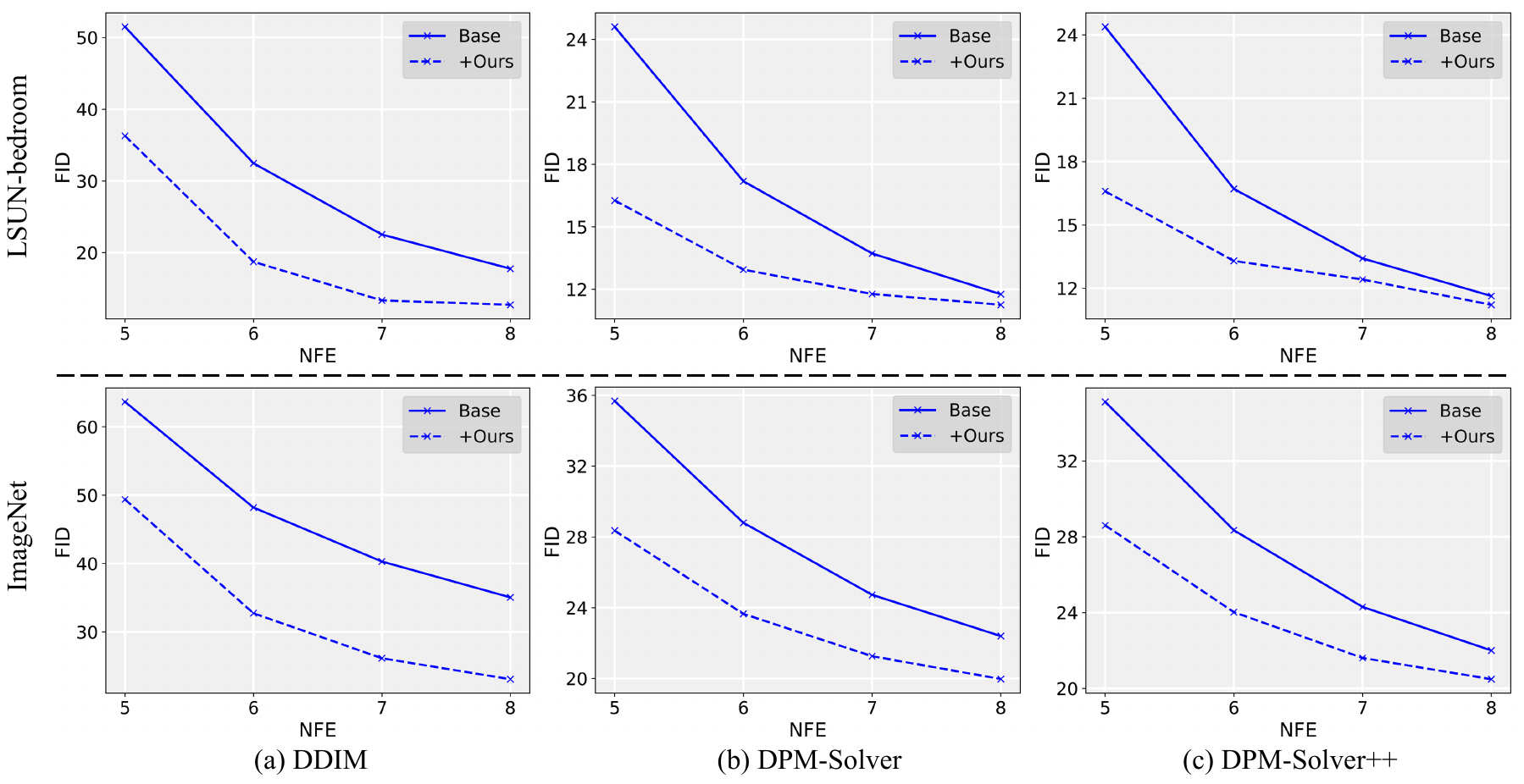}
        \caption{Unconditional sampling results. We compare SynBoost with baseline samplers on LSUN Bedroom and ImageNet datasets. We report the FID $\downarrow$ of the methods with different NFEs. Experimental results show that SynBoost is consistently better than baselines on pixel-space DPMs.}
        \label{fig:unconditional}
    \end{minipage}
    
    \begin{minipage}[b]{0.98\textwidth}
        \includegraphics[width=0.98\linewidth]{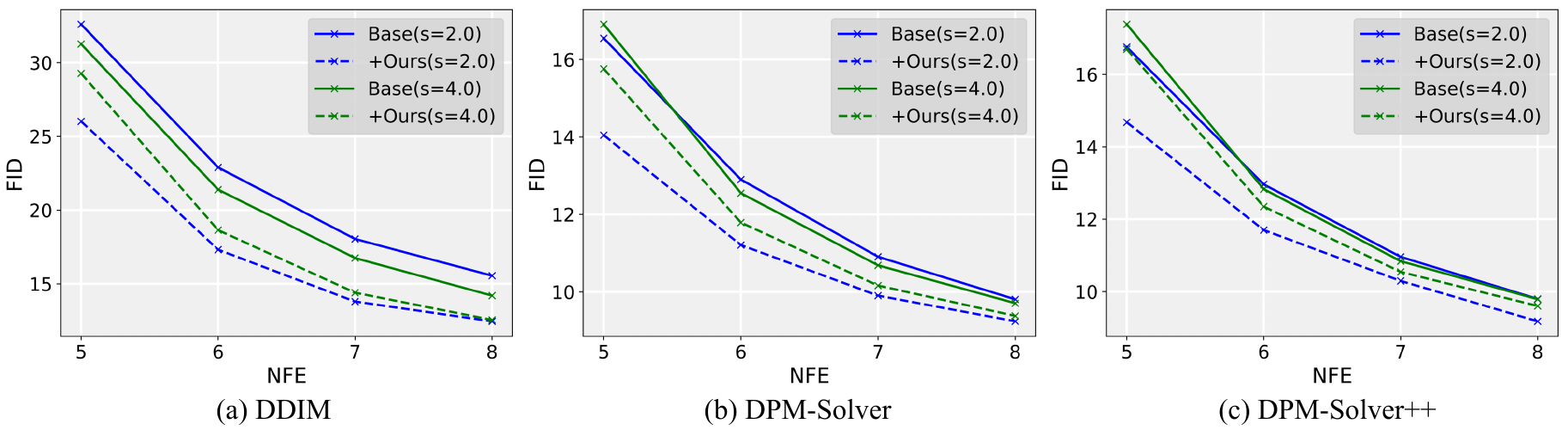}
        \caption{Class-conditional sampling results. Quantitative comparisons between existing samplers and our method with classifier-guided class condition, employing various classifier scales (s=2.0 and 4.0) and NFEs. SynBoost achieves consistent performance improvement over baseline samplers. } 
        \label{fig:class-conditional}
    \end{minipage}
    
    \begin{minipage}[b]{0.98\textwidth}
        \includegraphics[width=0.98\linewidth]{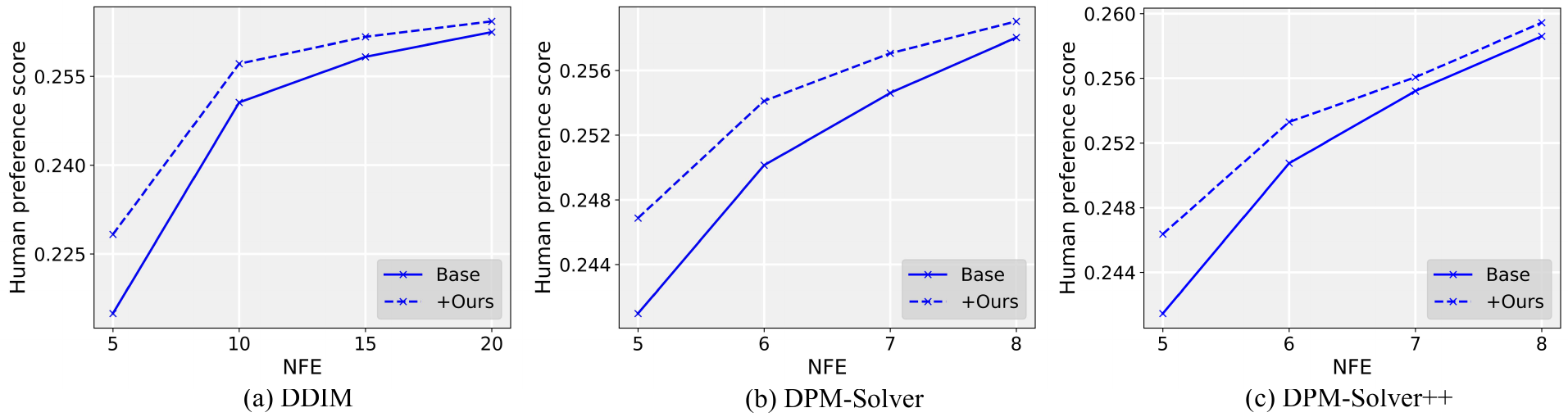}
        \caption{Text-conditional sampling results. The results are reported on Stable diffusion model with the MS-COCO2014 validation dataset and guidance scale 7.5. We adopt the human preference score $\uparrow$ from HPD v2 \cite{wu2023human}. SynBoost performs better than previous baselines on latent-space DPMs.}
        \label{fig:text}
    \end{minipage}

\end{figure*}

\subsection{Main Results}
\label{subsec:main}
We illustrate the effectiveness of our method on few-step sampling with comprehensive experiments and analyses. Our experiments cover the main fast ODE samplers with different orders (1-order DDIM, 2-order DPM-Solver and DPM-Solver++), two model prediction modes (noise and date prediction), three general generation types (unconditional, class-conditional, and text-conditional), two existing guiding strategies (classifier-guided \cite{dhariwal2021diffusion} and classifier-free guided \cite{ho2022classifier} models), two mainstream sampling space (image space \cite{dhariwal2021diffusion} and latent space \cite{rombach2022high}), main-stream datasets (LSUN-bedroom and ImageNet), as well as various guidance scales.

\noindent \textbf{Unconditional sampling.}
We first compare the unconditional sampling quality of different methods on LSUN Bedroom \cite{yu2015lsun} and ImageNet \cite{deng2009imagenet} datasets in Figure \ref{fig:unconditional}. The pre-trained diffusion models are from \cite{dhariwal2021diffusion}. Our method substantially boosts existing fast ODE solvers with better sampling quality and faster speed. The performance lift is especially obvious with fewer NFEs, which demonstrates the potential and effectiveness of SynBoost for the practical deployment of generative diffusion models. 

\noindent \textbf{Class-conditional sampling.}
Besides the unconditional sampling, we adopt the class label as condition information. To this end, we employ the classifier-guided sampling strategy and the pre-trained models from \cite{dhariwal2021diffusion}. We validate the effectiveness of our method over baseline samplers under different guidance scales (s = 2.0 and 4.0). The results are shown in Figure \ref{fig:class-conditional}. SynBoost achieves consistent performance improvement with various solvers and guidance scales.

\begin{figure*}[p]
    \centering
    \begin{minipage}[b]{0.99\textwidth}
        \includegraphics[width=0.99\linewidth]{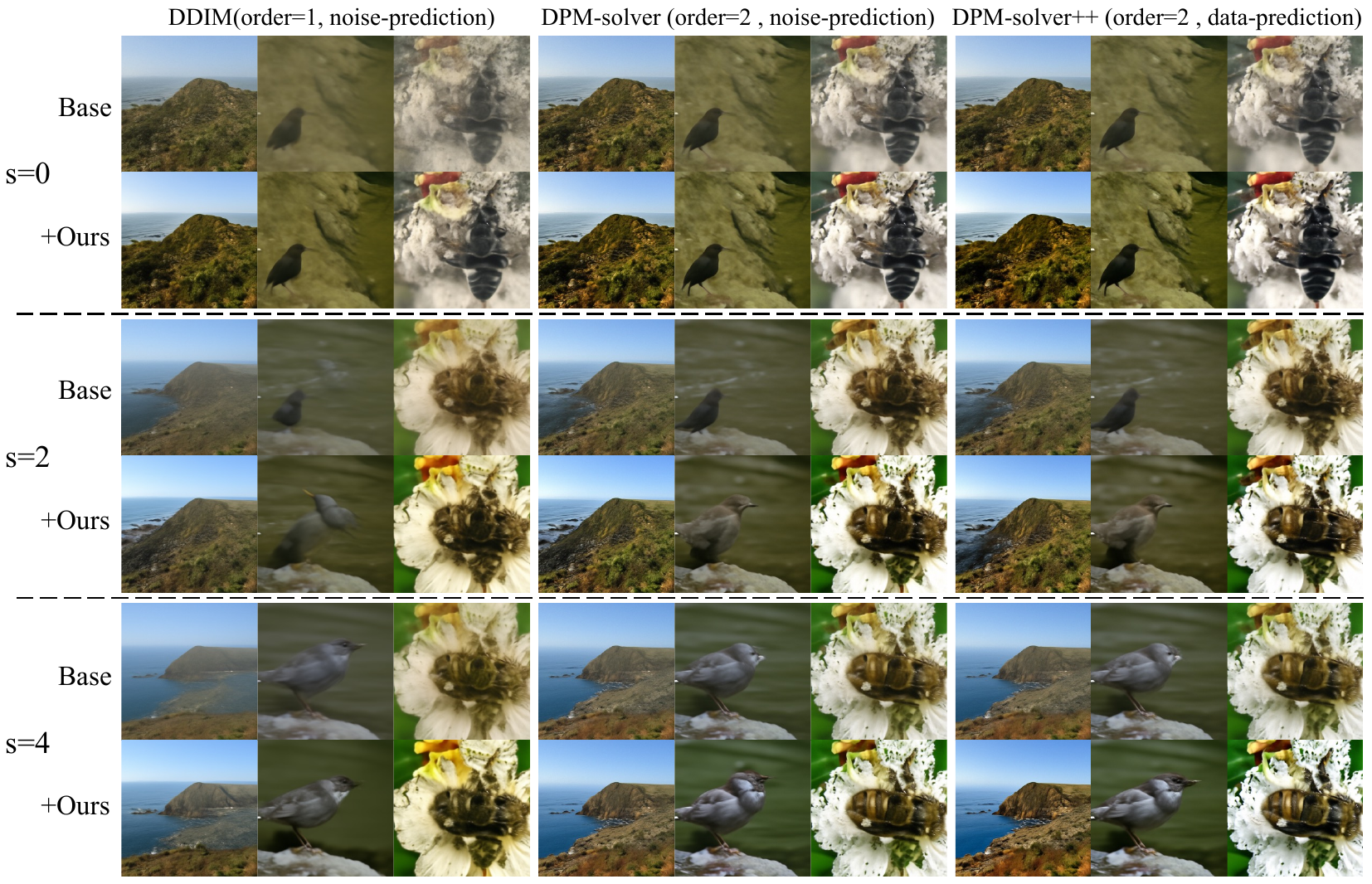}
        \setlength{\abovecaptionskip}{-0.02cm}
        \setlength{\belowcaptionskip}{0.1cm}
        \caption{Qualitative comparisons with class-conditional sampling in pixel-space. All images are generated by sampling from a DPM trained on ImageNet 256×256 \cite{dhariwal2021diffusion} with NFE= 7. The classifier scale $s$ is respectively set as 0.0, 2.0, and 4.0. Our method can generate more plausible samples with more visual details and higher contrast compared with previous samplers.}
        \label{fig:visual_pixel}
    \end{minipage}
    
    \begin{minipage}[b]{0.99\textwidth}
        \includegraphics[width=0.99\linewidth]{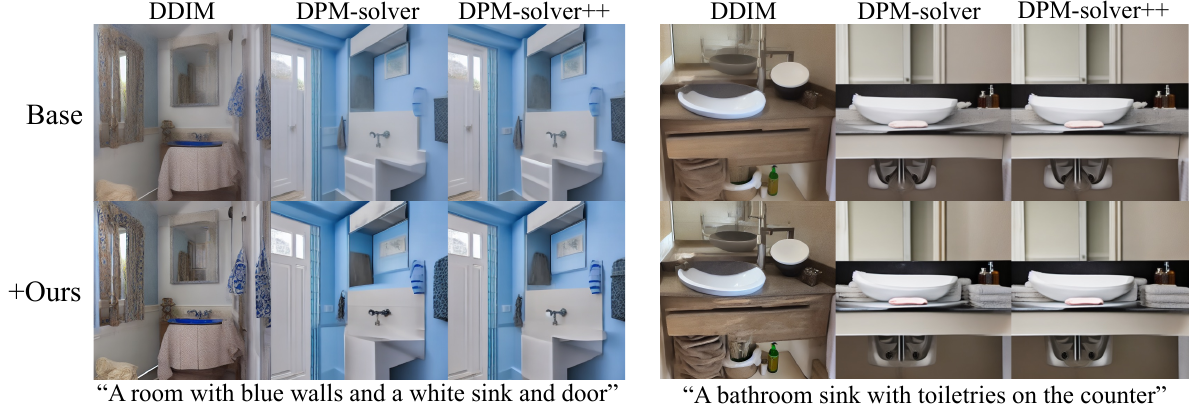}
        \setlength{\abovecaptionskip}{-0.02cm}
        \caption{Qualitative comparisons with text-conditional sampling in latent-space. All images are generated with NFE = 5 and classifier-free guidance scale = 7.5. Our method can consistently generate more realistic images with fewer visual flaws than previous samplers of various orders.}
        \label{fig:visual_latent}
    \end{minipage}

\end{figure*}


\begin{figure*}[t]
    \begin{center}
	\includegraphics[width=0.95\linewidth]{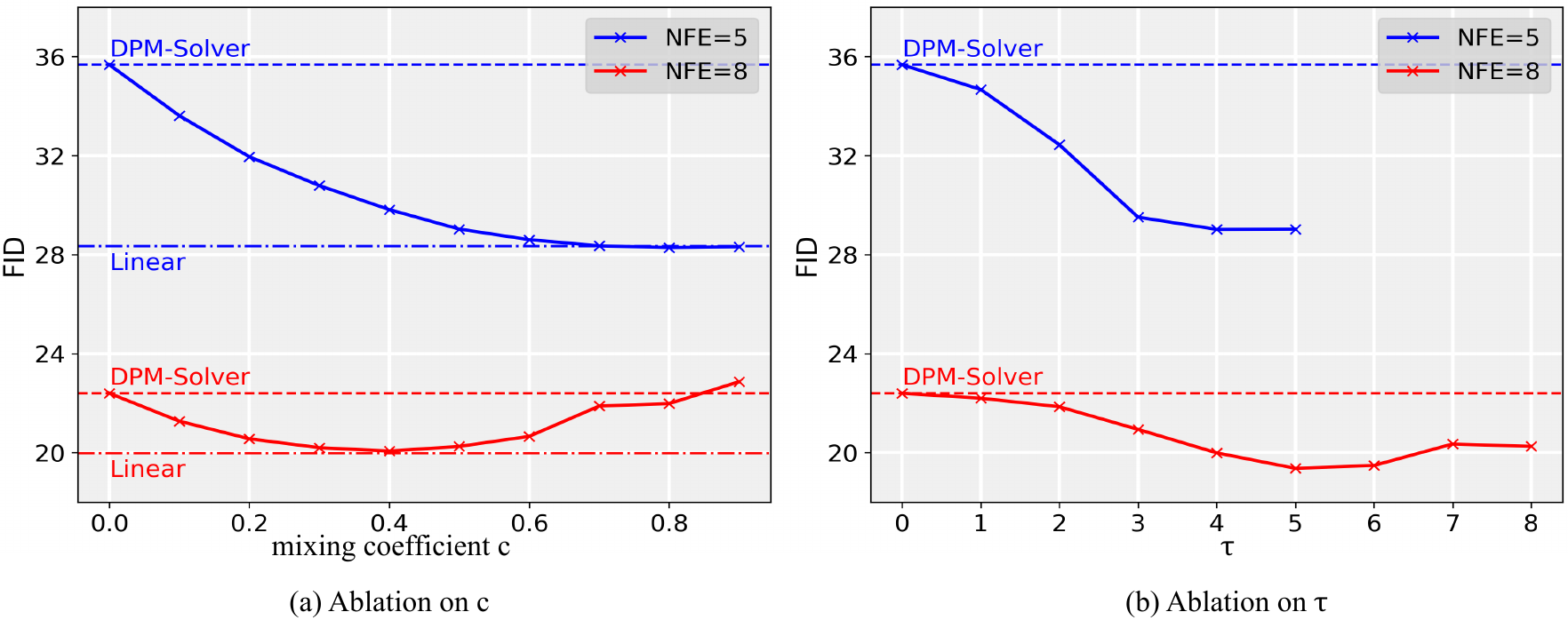}
    \end{center}
    \caption{Sensitivity analysis and robustness of the framework's hyper-parameters. This figure investigates the two primary components of Equation 9: (a) the mixing coefficient $c$ and (b) the target timestep $\tau$. 
    The results in (a) is achieved with $\tau$ fixed to $T$, showing that linearly decreasing strategy for $c$ outperforms constant coefficients.
    Part (b) demonstrates that, under the linear mixing strategy, choosing a larger $\tau$ generally yields better performance. Overall, the curves reflect high robustness, as a wide range of parameter choices consistently result in a substantial performance lift over the base DPM-Solver.}
    \label{fig:ablation}
\end{figure*}

\noindent \textbf{Ablations on $c$ and $\tau$.}
Figure \ref{fig:ablation} shows the sensitivity analysis and robustness of the framework's hyper-parameters, which investigates the two primary components of Equation 9: (a) the mixing coefficient $c$ and (b) the target timestep $\tau$. The results in (a) show that a linearly decreasing strategy for $c$ (starting at 0.5 and reaching 0.0 at step $T$) outperforms constant coefficients, aligning with the empirical observation that approximation error decreases as steps advance. Part (b) demonstrates that choosing a larger $\tau$ (closer to initial noise $T$) generally yields better performance due to the higher accuracy of the network at those timesteps. Overall, the curves reflect high robustness, as a wide range of parameter choices consistently result in a substantial performance lift over the base DPM-Solver.

\noindent \textbf{Text-conditional sampling.}
To further assess our method's performance across different condition types, we explore its application in a text-to-image stable diffusion model \cite{rombach2022high}, which works in latent space utilizing classifier-free guidance strategy \cite{ho2022classifier}. The guidance scale is set as 7.5 following the common setting. \revII{We concatenate conditional and unconditional inputs channel-wise into a single forward pass, consuming exactly 1 NFE per step.} We sample the first 10K captions from the MS-COCO2014 validation dataset \cite{lin2014microsoft} for input texts. Acknowledging the limitations of the FID metric in text-to-image scenarios \cite{lu2022dpm++, zhao2023unipc}, and the inadequacy of MSE for evaluating distribution convergence, we instead employ the HPD v2 \cite{wu2023human}, a state-of-the-art model that predicts human preferences for images generated by text-to-image diffusion models. Results, illustrated in Figure \ref{fig:text}, show our method consistently outperforms baseline samplers across varying orders, as evidenced by higher human preference scores.

\subsection{Visual Results}
\label{subsec:visual}
\noindent \textbf{Class-conditional sampling.}
We provide a qualitative comparison between our method and previous sampling methods in Figure \ref{fig:visual_pixel}. We adopt NFE=7 with various classifier scales ($s$ = 0.0, 2.0, and 4.0) for the presented samples. 
We find that the sampling quality generally increases as the order of the sampler increases under the same NFE. Besides, the class labels of the images are increasingly obvious due to larger classifier guidance scale. While, our method consistently improves the image quality with better details, color, and contrast regardless of the sampler and guidance scale. For example, DualFast can even boost DDIM to achieve comparable visual results to DPM-Solver++. The above results strongly demonstrate the effectiveness of our method on pixel-space based sampling.
Our method consistently improves the image quality with better details, color, and contrast regardless of the samplers and guidance scales. For example, DualFast can even boost DDIM to achieve comparable visual results to DPM-Solver++. The above results strongly demonstrate the effectiveness of our method.

\noindent \textbf{Text-conditional sampling.}
Besides the pixel-space sampling results, we additionally provide visual results on stable diffusion in Figure \ref{fig:visual_latent}. The NFE is set as only 5 to validate the performance bound of the compared methods. The classifier-free guidance scale is 7.5. Our method consistently generate more realistic images with fewer visual flaws and better structures than previous samplers. The above results illustrate that our method generalizes well to both pixel and latent space. 
Besides the pixel-space sampling results, we additionally provide a qualitative comparison between our method and previous sampling methods on stable diffusion model in Figure \ref{fig:visual_latent}. The NFE is set as only 5 to validate the performance bound of existing sampling methods and our method. The classifier-free guidance scale is 7.5. Our method consistently generate more realistic images with fewer visual flaws and better structures than previous samplers of various orders. The above results illustrate that our method generalizes well to both pixel and latent space generation.

\subsection{Ours-UniPC}  \label{subsec:Ours_UniPC}
UniPC is the recent state-of-the-art high-order ODE solver. It employs a predictor-corrector framework, consisting of a predictor and a corrector. The unified corrector (UniC) can be applied after any existing DPM sampler to increase the order of accuracy without extra model evaluations, and the unified predictor (UniP) supports arbitrary order. In this part, we take the most widely used third-order UniPC sampler as example. Concretely, the predictor in UniPC first gets an estimation of $\x_{t}^p$, then with the corresponding $\D_{t-1}$: 
\begin{equation}
    \label{supeq:1}
    \begin{aligned}
    \D_{t-1}^{base} &= \epsilonv_{\theta}(\x_{t}, t) + \sum_{m=0}^{p-2} a_m^p  [ \epsilonv_{\theta}(\x_{t+m}, t+m) \\
    &- \epsilonv_{\theta}(\x_{t+1+m}, t+1+m) ], \\
    \end{aligned}
\end{equation}
where $p$ is the order of the predictor. Then the corrector in UniPC refines the estimation $\x_{t}^p$ with the corresponding $\D_{t}$: 
\begin{equation}
        \label{supeq:2}
        \begin{aligned}
        \D_{t-1}^{\text {base}} & =\epsilonv_{\theta}(\x_{t}, t)+\sum_{m=0}^{p-2} a_{m}^c [ \epsilonv_{\theta}(\x_{t+m}, t+m) \\
        & - \epsilonv_{\theta}(\x_{t+1+m}, t+1+m) ] \\
        & +a_{0}^c\left[\epsilonv_{\theta}(\x_{t-1}, t-1) - \epsilonv_{\theta}(\x_{t}, t)\right].
        \end{aligned}
\end{equation} 

Then the modified version with our approximation error reduction strategy are presented as follows:
\begin{equation}
        \label{supeq:3}
        \begin{aligned}
        \D_{t-1}^{\text {base}} & = \epsilonv_{\theta}(\x_{t}, t) + \sum_{m=0}^{p-2} a_{m}^c [ \epsilonv_{\theta}(\x_{t+m}, t+m) \\
        & - \epsilonv_{\theta}(\x_{t+1+m}, t+1+m) ] +a_{0}^c[(1+c)\epsilonv_{\theta}(\x_{t-1}, t-1) \\
        &- c \epsilonv_{\theta}(\x_{\tau}, \tau ) - \epsilonv_{\theta}(\x_{t},t)].
        \end{aligned}
\end{equation}

\begin{figure*}[htbp]
    \begin{center}
	\includegraphics[width=0.9\linewidth]{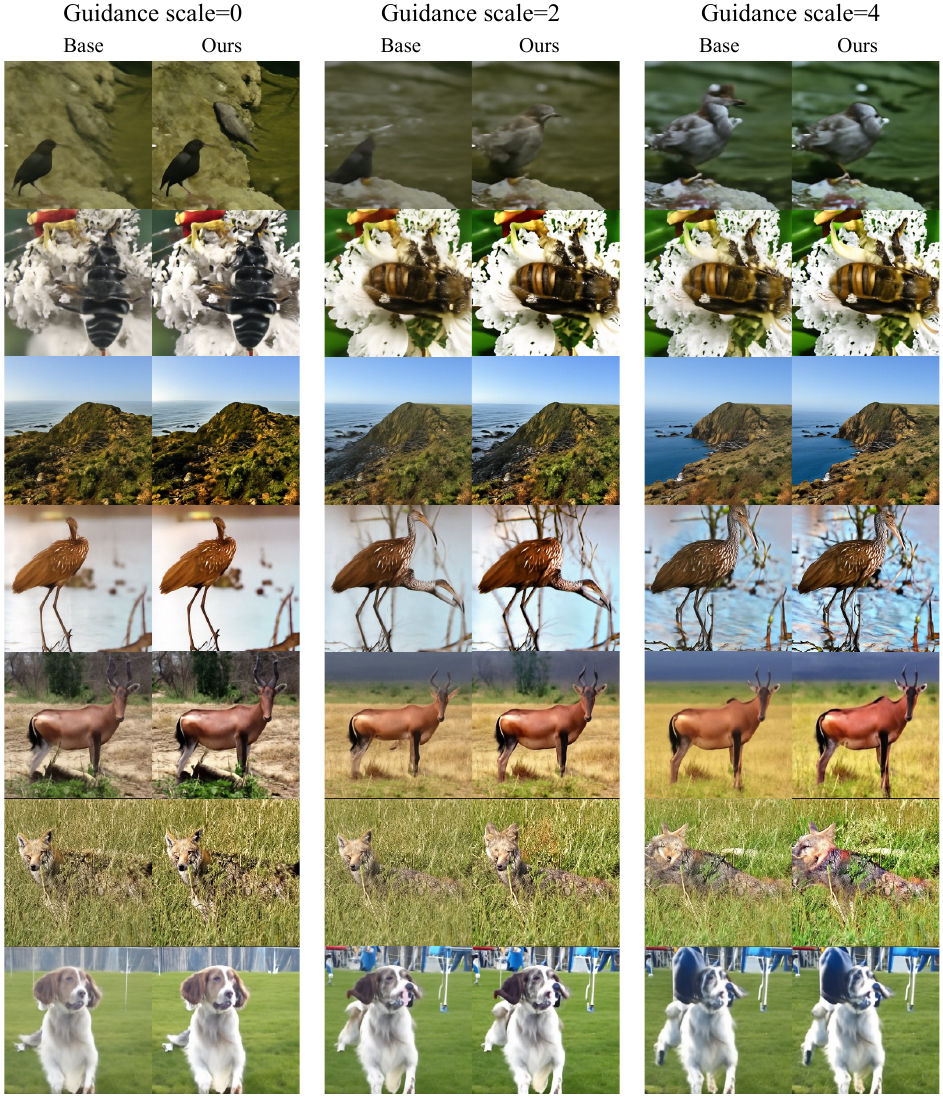}
    \end{center}
    \caption{Qualitative comparison of SynBoost integration with UniPC sampler on ImageNet dataset with various guidance scales. These images provide a visual validation of the quantitative data in Table \ref{tab:unipc}. Our method exhibits noticeably better structural details, higher color contrast, and fewer artifacts compared to the base UniPC sampler. This demonstrates that by correcting the approximation bias at each step, SynBoost enables high-order solvers to produce more realistic and plausible samples in the pixel-space, even when the underlying solver is already highly optimized for discretization errors.
    } 
    \label{fig:sup_unipc_pixel}
\end{figure*}

We also show the results on UniPC sampler with quantitative comparison in Table \ref{tab:unipc} and visual results in Figure \ref{fig:sup_unipc_pixel}. Our method can also boost the performance of this 3-order solver.

\begin{table*}[ht]
    \setlength{\tabcolsep}{12pt}
    \centering
    \caption{Reduced error. Detailed MSE reduction analysis between DPM-Solver and SynBoost at various NFEs. By using the results of 1000-step sampling as pseudo-ground truth, the experiment demonstrates that SynBoost consistently achieves lower MSE than the base DPM-Solver across all steps. Since the discretization error remains identical for both methods at a given NFE, the observed reduction stems directly from SynBoost’s ability to mitigate the ``approximation error". }
    \begin{tabular}{c|ccccccccc}
    \hline 
    \multirow{2}{*}{MSE($10^{-3}$)}     & \multicolumn{8}{c}{NFE}     \\   \cline{2-10} 
	      & 5 & 6 & 7 & 8 & 9 & 10 & 12 & 15 & 20                 \\ \hline
	Base    & 10.97 & 8.19 & 6.23 & 4.93 & 3.96 & 2.63 & 1.82 & 1.11 & 0.61   \\
	  Ours    & 7.81  & 5.28 & 3.69 & 2.79 & 2.15 & 2.08 & 1.44 & 0.96 & 0.53   \\ \hline
    \end{tabular}
    \label{table:MSE}
\end{table*}

\definecolor{lightgray}{gray}{0.5}
\setlength{\arrayrulewidth}{0.5pt} 
\begin{table*}[htbp]
\setlength{\tabcolsep}{16pt}
\caption{Performance enhancement of unconditional sampling on the high-order UniPC solver. This table quantifies the improvement in FID scores when the SynBoost framework is integrated into UniPC, a high-order predictor-corrector ODE solver. While UniPC is designed to minimize discretization error through its multi-step architecture, the further reduction in FID achieved by ``UniPC (Ours)" proves that approximation error remains a bottleneck even for advanced third-order solvers.} 
\label{tab:unipc}
\centering
\begin{tabular}{l|cccc}
	\toprule
        Sampling Method $\backslash$ NFE  & 4 & 5 & 6 & 7  \\  
        \hline
	  UniPC (Base)  & 44.617  & 29.472  & 23.324  & 20.676           \\
	  UniPC (Ours)  & 42.066  & 28.367  & 22.882  & 20.451           \\
        \bottomrule
\end{tabular}
\end{table*}

\subsection{Analyses}  \label{subsec:analysis}
In this section, we will provide more detailed analyses and ablation studies (including: the choices of hyper-parameters, MSE analysis, higher guidance scale, larger NFEs, and compatibility with DiT \cite{peebles2023scalable} architecture) to further evaluate the effectiveness of SynBoost. Due to page limit, \textbf{we leave more experiments and analyses in the supplementary material, including} the  comparison with SDE-based samplers, performance upper bound of SynBoost, sampling diversity metric, as well as more visual results.

\noindent \textbf{Ablation on the choices of $c$ and $\tau$.}
SynBoost has two main hyper-parameters within equation \ref{eq:8}. For the mixing coefficient $c$, based on the prior that approximation error linearly decreases with steps, we adopt a linearly decreasing strategy (from 0.5 to 0.0), which starts from 0.5 at step 0 and reaches 0.0 at step T. We also compare with constant mixing coefficient in Figure \ref{fig:ablation}, where our linearly decreasing weight achieves higher performance than constant one. For the choice of step $\tau$, we investigate the performance of different $\tau$ in Figure \ref{fig:ablation}. Higher $\tau$ usually corresponds to better performance. For simplicity and ease-of-use, we directly employ $\tau=T$. This means that we employ initial noise $\x_T$ as $\epsilonv_{\theta}(\x_{\tau}, \tau)$.
Besides the above analyses, we additionally emphasise that due to our efficient paradigm, the choices of $c$ and $\tau$ are quite robust. Different choices all lead to substantial performance lift compared to the baseline solver.

\begin{figure}[ht]
    \begin{center}
    \includegraphics[width=0.90\linewidth]{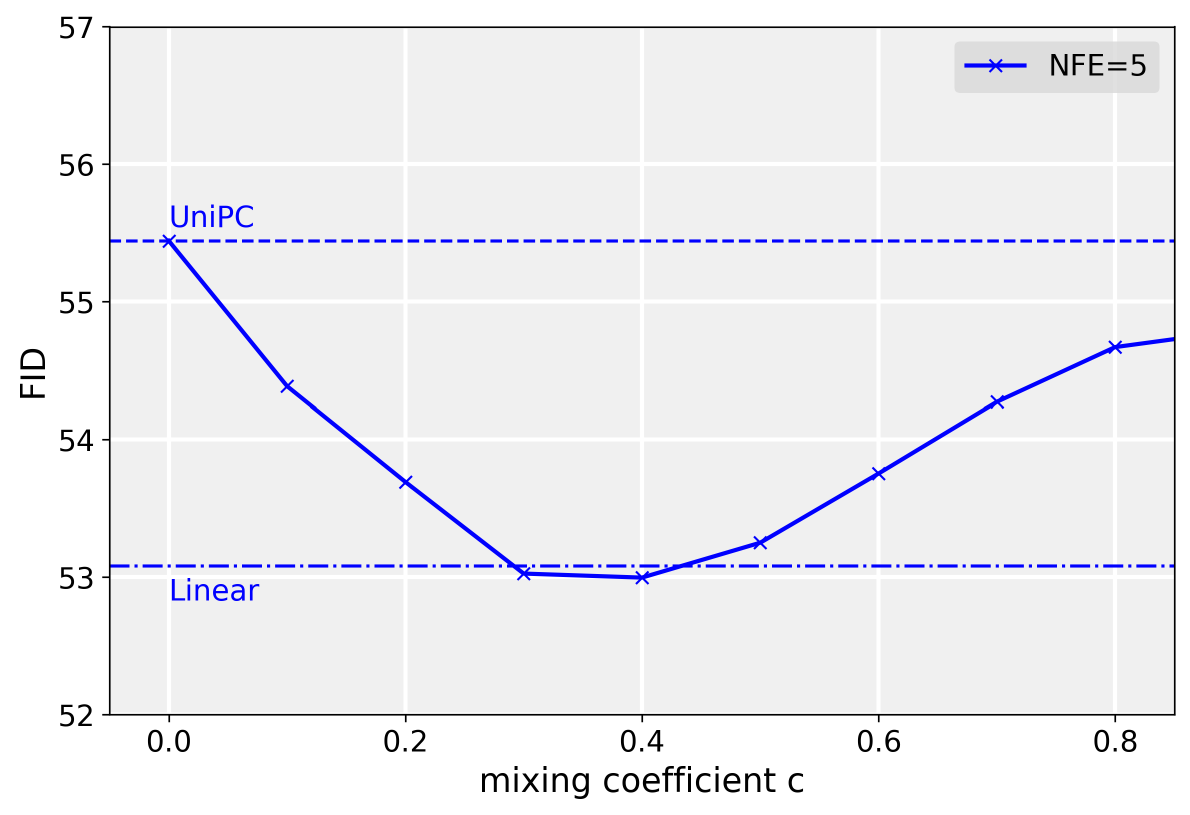}
    \end{center}
    \caption{Generality of the mixing strategy across different solver: ablation study of the mixing coefficients c on the UniPC sampler using $NFE=5$. The U-shaped FID curve confirms that an optimal mixing coefficient $c$ (around 0.3 to 0.4) significantly improves generation quality compared to the base solver ($c=0.0$). The consistency of this trend with the results for 2nd-order solvers in Figure \ref{fig:ablation} reinforces the generality of the SynBoost framework. It proves that the linear mixing strategy remains a robust and effective method for reducing total sampling error regardless of the mathematical order or complexity of the base ODE solver.
    } 
    \label{fig:rebuttal_mixing}
\end{figure}

We also expanded the ablation study on the mixing coefficient $c$ to include the 3rd-order UniPC solver in Figure \ref{fig:rebuttal_mixing}. The experiment is conducted on the ImageNet dataset with unconditional generation. We adopt NFE=5 for lower computational burden. The results confirm that our linear mixing strategy for $c$ remains highly effective and robust even when integrated with higher-order solvers like UniPC, further demonstrating the generality of SynBoost. 

\noindent \textbf{Reduced error.}
We verify the effectiveness of SynBoost with the common FID metric as well as numerical visual results. 
Besides, we also depict the MSE analysis in Table \ref{table:MSE}. Specifically, we adopt the results of 1000-step sampling as pseudo GT. Then we generate samples under various NFEs with both baseline solver and our SynBoost. Since the NFEs and sampler orders are kept identical between the baseline solver and our SynBoost, the discretization errors are also same between these two pairs. Thus the additional MSE error reduction stems from  the decreased approximation error brought by our SynBoost. More concretely, at each step $t$, the network prediction is modified with higher accuracy, thus contributing to smaller final error compared to the pseudo GT.

\noindent \textbf{Higher Guidance Scale.}
Guided sampling can significantly boost the sample quality compared to unconditional sampling. But high guidance scale would also cause the instability of the sampler \cite{lu2022dpm++} and poor sample quality. In this part, we show the results with guidance scale of 6.0 in Table \ref{tab:guidance_scale}, and avoid higher guidance scale. Compared with the results of smaller guidance scales, the sample quality is worse with guidance scale 6.0. But, SynBoost can still consistently achieve better performance than the base sampler.

\definecolor{lightgray}{gray}{0.5}
\setlength{\arrayrulewidth}{0.5pt} 
\begin{table*}[htbp]
\setlength{\tabcolsep}{14pt}
\caption{Robustness and stability under high classifier guidance ($s=6.0$) of class-conditional sampling. High guidance scales often lead to sampler instability and degraded sample quality in standard diffusion models. However, the data shows that SynBoost consistently outperforms the base DDIM (e.g., reducing FID from 23.408 to 19.845 at NFE=6). This demonstrates that SynBoost effectively stabilizes the sampling trajectory even when aggressive guidance is applied.} 
\label{tab:guidance_scale}
\centering
\begin{tabular}{c|l|cccc}
	\toprule
        Guidance Scale & Sampling Method $\backslash$ NFE  & 5 & 6 & 7 & 8   \\  
        \midrule
	\multirow{2}{*}{6.0}   & DDIM (Base)  & 36.129  & 23.408  & 17.407  & 14.705           \\
	  & DDIM (Ours)  & 34.976  & \revII{19.845}  & \revII{14.341}  &  12.740                 \\
        \bottomrule
\end{tabular}
\end{table*}

\begin{figure}[ht]
    \begin{center}
	\includegraphics[width=0.95\linewidth]{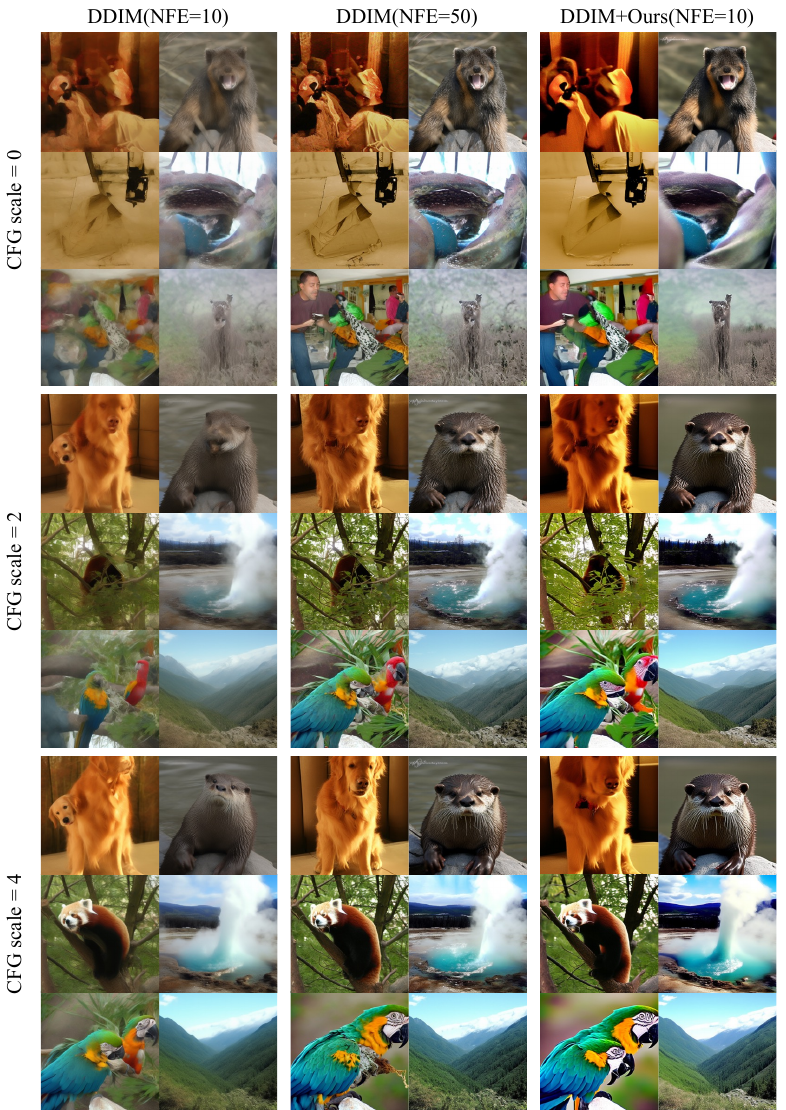}
    \end{center}
    \caption{Visual results on DiT with various guidance scales. All images are generated by sampling from DiT-XL/2 \cite{peebles2023scalable} with class condition. Our method significantly boosts the quality and speed of DDIM sampler, even achieving comparable visual results to DDIM of 50 NFEs with only 10 NFEs. } 
    \label{fig:DiT}
\end{figure}

\noindent \textbf{Performance on DiT.}
We present the efficacy of our method on DiT \cite{peebles2023scalable}, which adopts transformer \cite{vaswani2017attention} architecture. Specifically, we adopt DiT-XL/2 with various guidance scales in Figure \ref{fig:DiT}. Our method significantly boosts the quality and speed of DDIM sampler, even achieving comparable visual results to DDIM of 50 NFEs with only 10 NFEs.

\noindent \textbf{Larger NFEs.}
In this part, we also validate the effectiveness of our method on larger NFEs in Table \ref{tab:NFEs}. SynBoost improves the FID of DDIM from 28.906 to 20.559 when NFE=10, achieving two times acceleration (comparable to 20-step DDIM sampling).

\definecolor{lightgray}{gray}{0.5}
\setlength{\arrayrulewidth}{0.5pt}
\begin{table*}[ht]
\setlength{\tabcolsep}{10pt}
\caption{Sample quality of unconditional sampling on ImageNet dataset with larger NFEs. This table explores the effectiveness of SynBoost as the number of sampling steps increases from 10 to 20. A pivotal finding is that DDIM (Ours) at only 10 NFEs (FID 20.559) achieves comparable or superior results to the base DDIM at 20 NFEs (FID 20.344). This indicates a roughly 2x acceleration in inference speed while maintaining image quality.
} 
\label{tab:NFEs}
\centering
\begin{tabular}{c|ccc|ccc|c|c}
	\hline
        Sampler &  \multicolumn{3}{c|}{DDIM (Base)}   &  \multicolumn{3}{c|}{DPM-Solver (Base)}   & DDIM (Ours) & \revIII{DPM-Solver (Ours)} \\
        \hline
        NFE     & 10 & 15 & 20    & 10 & 15 & 20    & 10   & \revIII{10} \\  
        \hline
	FID   & 28.906  & 22.781  & 20.344  & 21.250  & 20.004  & 19.533   & 20.559   & \revIII{19.727}   \\
        \hline
\end{tabular}
\end{table*}

\begin{figure}[htbp]
    \begin{center}
    \includegraphics[width=0.96\linewidth]{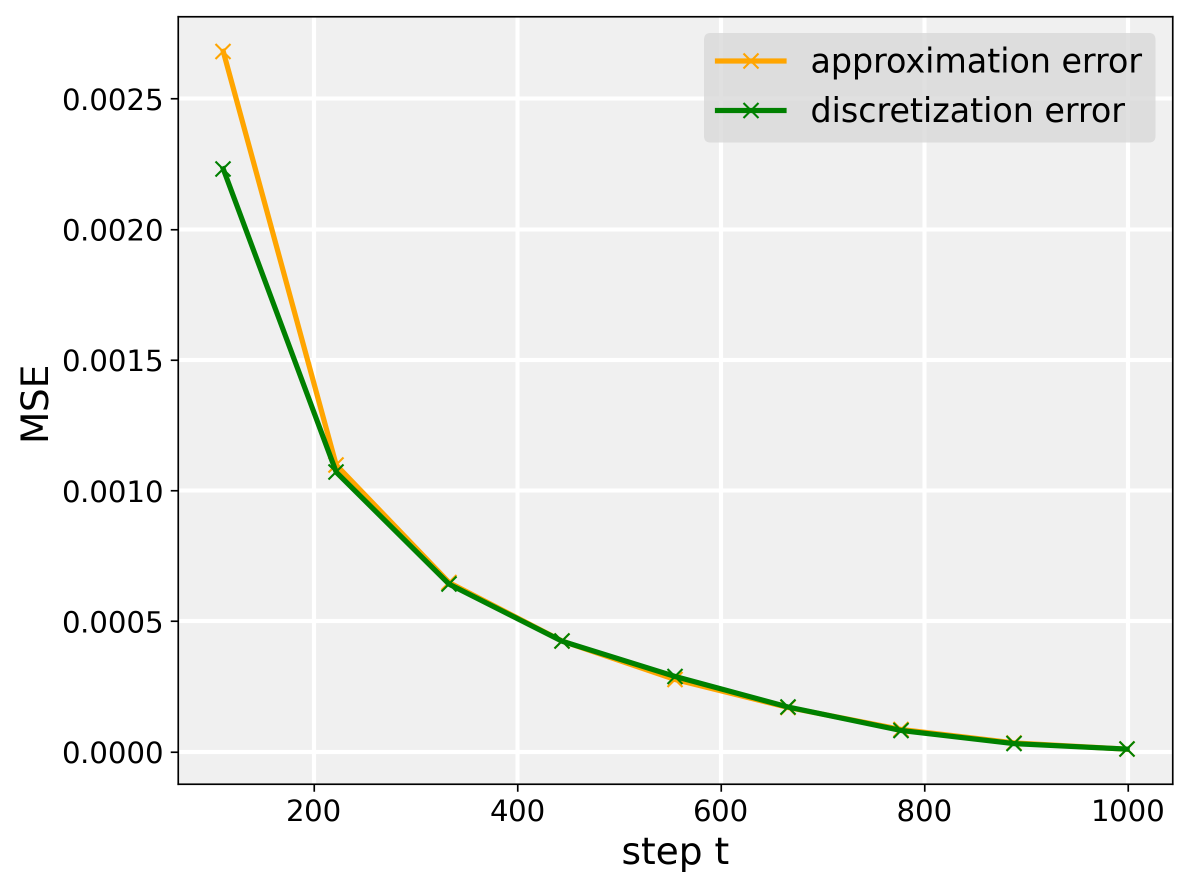}
    \end{center}
    \caption{Empirical verification of error trends in guided sampling tasks. To ensure the observations from Figure \ref{fig:error} are not limited to unconditional task, this figure illustrates the error disentanglement for classifier-guided generation on ImageNet. The curve confirms the universal monotonic trend that approximation error consistently decreases as the timestep $t$ increases toward $T$. This consistency across different generation settings provides the fundamental theoretical grounding for the SynBoost design.} 
    \label{fig:rebuttal_error}
\end{figure}

\noindent \textbf{Extension of the observed trend.}
The statistical results in Figure \ref{fig:error} are evaluated on the unconditional generation task using the ImageNet dataset. To verify the consistency of this observation on different generation settings, we present the error decomposition curve for classifier-guided class-conditional generation on ImageNet in Figure \ref{fig:rebuttal_error}. The results exhibit the exact same monotonic trend that approximation error decreasing monotonically as timestep $t$ increases.

\noindent \textbf{MSE for distributions measurement in diffusion models.}
Adopting MSE to measure the distributions divergence in diffusion models is grounded with both theoretical guarantee and sufficient empirical support from classical and representative papers.
(1) Employing MSE to measure the distribution divergence in this special case is theoretically guaranteed. \cite{box2011bayesian} discusses MSE as a special case of maximum likelihood estimation when the error follows a Gaussian distribution. 
In the context of deep learning, \cite{lecun2015deep} discusses the application of MSE, particularly in error measurement in generative models. 
Since in the context of diffusion models, gaussian distribution is the essential and default choice. MSE is thus a rational metric to measure the distribution divergence.
(2) It is also a common practice of previous sampler pappers, that employing MSE to measure distribution distance. For example, the main-stream samplers (our baselines), including DPM-Solver++ \cite{lu2022dpm++} and UniPC \cite{zhao2023unipc}, also employs MSE (l2 distance) to compare the convergence error between the results of different methods and 1000-step DDIM. Besides, EDM \cite{karras2022elucidating} focuses on the discretization error and also proposes to leverage root mean square error (RMSE) to measure the distribution distance between one Euler iteration and a sequence of multiple smaller Euler iterations, representing the ground truth.

\noindent \textbf{Discussions and limitations.}
Besides, despite the effectiveness of SynBoost, it still lags behind training-based methods \cite{salimans2022progressive, song2023consistency} with one-step generation. How to further close the gap between training-free methods and training-based methods requires future efforts.

\section{Conclusion}
In this work, we re-examine the fundamental sampling process of diffusion probabilistic models and reveal that the total sampling error is composed of two distinct parts: the well-studied discretization error and the under-explored approximation error. While existing fast samplers primarily focus on mitigating discretization error through high-order Taylor expansions, our dual-error disentanglement strategy elucidates that approximation error, arising from the neural network's imprecise estimation of the ground-truth score function, is of a comparable magnitude and significantly impacts sampling quality. Building on the empirical discovery that approximation error monotonically decreases as the timestep $t$ increases, we propose SynBoost, a unified and training-free acceleration framework that simultaneously addresses both error sources. By blending current noise estimations with more accurate predictions from larger timesteps, SynBoost seamlessly integrates as a plug-and-play strategy with existing solvers like DDIM, DPM-Solver, and UniPC. We verify the effectiveness of our method through extensive experiments across unconditional and conditional generation tasks in both pixel-space and latent-space, demonstrating that SynBoost significantly elevates sample quality and efficiency, particularly in regimes with extremely limited sampling steps.
 
\bibliographystyle{IEEEtran}
\bibliography{reference}

\begin{IEEEbiography}
    [{\includegraphics[width=1in,height=1.25in,clip,keepaspectratio]{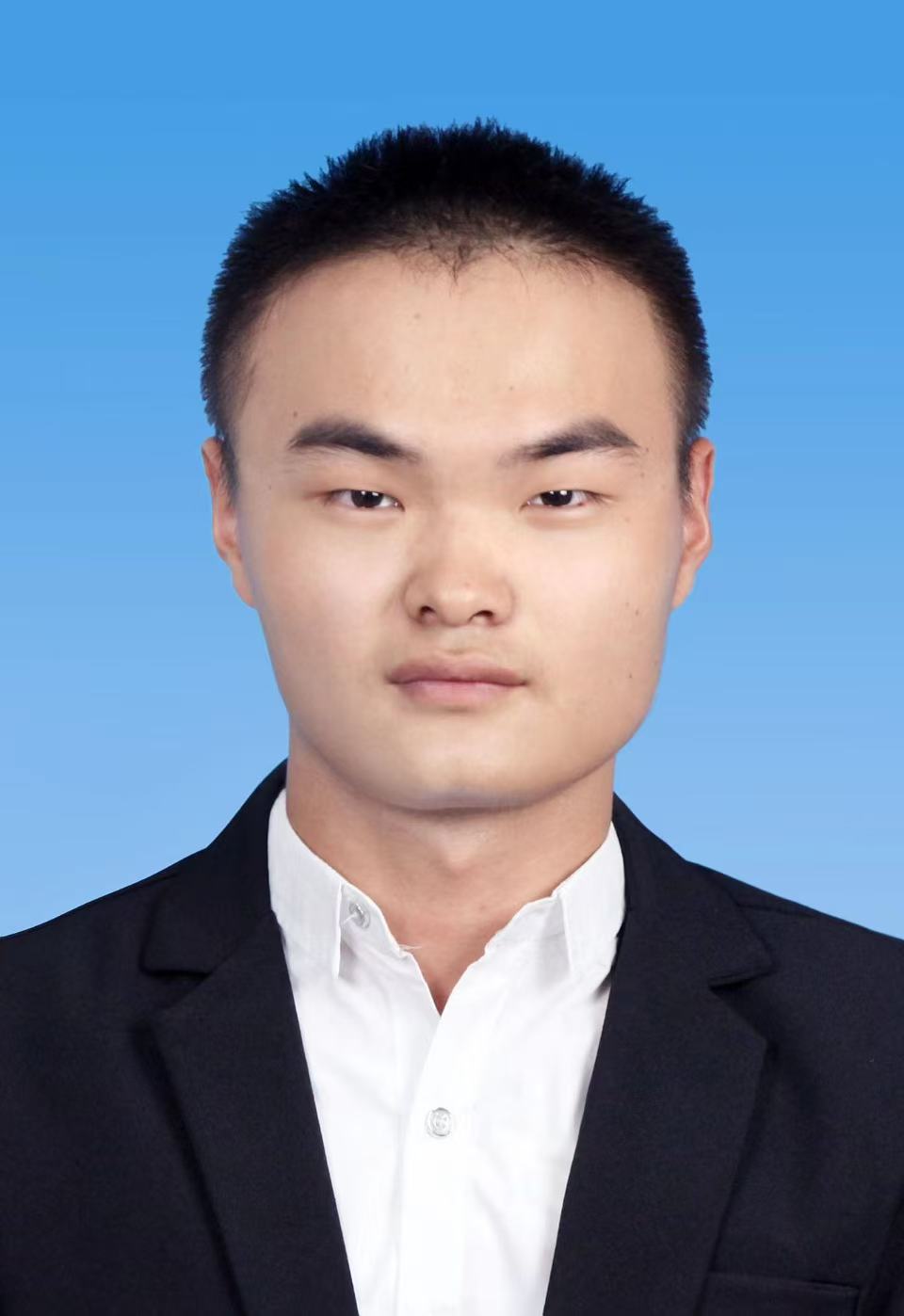}}]{Hu Yu} received the B.Eng. degree in electronic engineering from the University of Science and Technology of China (USTC), Hefei, China, in 2021.

    He is currently pursuing the Ph.D. degree with the University of Science and Technology of China (USTC), Hefei, China. His research interests include generative models, and representation learning.
\end{IEEEbiography}

\begin{IEEEbiography}
    [{\includegraphics[width=1in,height=1.25in,clip,keepaspectratio]{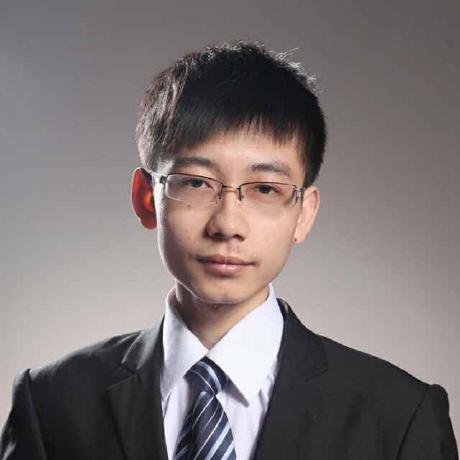}}]{Hao Luo} received the BS and PhD degrees from Zhejiang University, China, in 2015 and 2020, respectively. He is currently with the DAMO Academy, Alibaba Group and Hupan Lab, Zhejiang Province. His research interests include computer vision, vision transformer, generate models, and deep learning.
\end{IEEEbiography}

\begin{IEEEbiography}
    [{\includegraphics[width=1in,height=1.25in,clip,keepaspectratio]{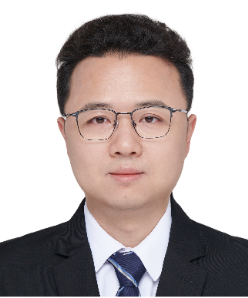}}]{Xueyang Fu} received the Ph.D. degree in signal and information processing from Xiamen University, Xiamen, China, in 2018.
    	
    He was a Visiting Scholar with Columbia University, New York, NY, USA, sponsored by the China Scholarship Council, from 2016 to 2017. He is currently a Professor and PhD Advisor with the Department of Automation, University of Science and Technology of China, Hefei, China. His research interests include machine learning and image processing.
\end{IEEEbiography}

\begin{IEEEbiography}
    [{\includegraphics[width=1in,height=1.25in,clip,keepaspectratio]{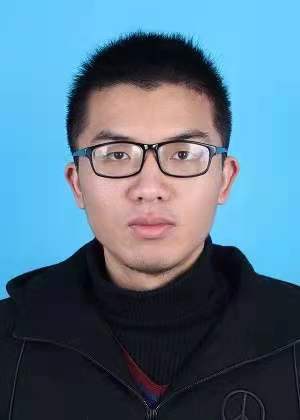}}]{Jie Huang} received the  Ph.D. degree in electronic engineering from the University of Science and Technology of China (USTC), Hefei, China, in 2024. He is currently an algorithm expert in Joy Future Academy, JD Group. His research interests include vision restoration and generation.
    
\end{IEEEbiography}

\begin{IEEEbiography}
    [{\includegraphics[width=1in,height=1.25in,clip,keepaspectratio]{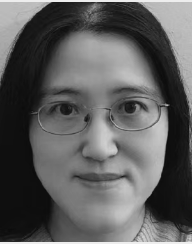}}]{Fan Wang} received the B.S. and M.S. degrees from the Department of Automation, Tsinghua University, Beijing, China, and the Ph.D. degree from the Department of Electrical Engineering, Stanford University, CA, USA. She is currently with the Alibaba Group as a Senior Staff Algorithm Engineer. Her research interests include object tracking and recognition, 3D vision, and multi-sensor fusion.
\end{IEEEbiography}

\begin{IEEEbiography}
    [{\includegraphics[width=1in,height=1.25in,clip,keepaspectratio]{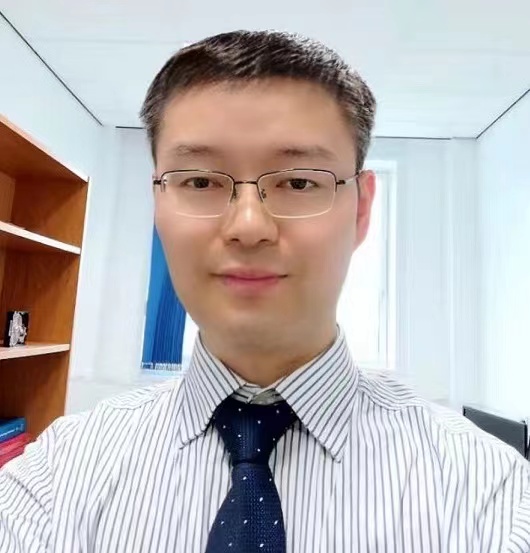}}]{Feng Zhao} (Member, IEEE) received the B.Eng. degree in electronic engineering from the University of Science and Technology of China (USTC), Hefei, China, in 2000, and the M.Phil. and Ph.D. degrees in computer vision from The Chinese University of Hong Kong, Hong Kong, in 2002 and 2006, respectively. From 2006 to 2018, he was a Post-Doctoral Fellow, Research Fellow, and permanent Senior Lecturer in Hong Kong, Singapore, and the United Kingdom. Since 2019, he has been a Professor and a Ph.D. Advisor with USTC. He has authored or coauthored more than 170 papers in referred international journals and conferences, including IEEE Transactions (e.g., TPAMI, TIP, TNNLS, TMM), Pattern Recognition, NeurIPS, ICML, ICLR, CVPR, ICCV, ECCV, IJCAI, AAAI, ACM MM, EMNLP, and ACL. His research interests include foundation model, AIGC, AI4Science, machine learning, computer vision, and image processing. Prof. Zhao was the recipients of the Best Paper Award in ICIDM 2018 and the Outstanding Paper Award in ACL 2024, the Champion of IJCAI-PRICAI 2020 3D AI Challenge, the Runner-up of CVPR 2021 Streaming Perception Challenge, and the Champions of ICCV 2023 Continual Test-time Adaptation for Semantic Segmentation Challenge and Multitask Learning for Robustness Challenge. He was the Program Committee Chairs of the International Conference on AI-Generated Content (AIGC) 2023-2025, and the Area Chairs of ACM MM 2023-2026, ACL 2025-2026, and AAAI 2026.
\end{IEEEbiography}

\clearpage
\newpage
\appendices

\renewcommand{\thesection}{S-\Alph{section}}
\renewcommand{\thesectiondis}{S-\Alph{section}}

\twocolumn[
  \begin{center}
    {\Huge \bfseries Supplementary Material for ``SynBoost: A Synergistic Framework for Fast Sampling of Diffusion Models'' \par}
    \vspace{1.5em}
  \end{center}
]

This supplementary document is organized as follows: 
\begin{itemize}
    \item Section~\ref{sec:quantitative} shows the quantitative comparisons.
    \item Section~\ref{sec:SDE} shows comparison with SDE-based sampler.
    \item Section~\ref{sec:bound} explores the performance upper bound of the baselines and SynBoost.
    \item Section~\ref{sec:diversity} shows the sampling diversity metric.
    \item \revI{Section~\ref{sec:explanations} presents more explanations of equation 9 in the manuscript.}
    \item Section~\ref{sec:Visual} depicts more visual results.
\end{itemize}

\vspace{1em}

\section{Quantitative comparison}    \label{sec:quantitative}

In the main manuscript, we show the quantitative comparison with NFE-FID curve. In this part, we additionally present all comparisons in Table \ref{tab:unconditional}, \ref{tab:class_conditional} and \ref{tab:text_conditional}. In Table \ref{tab:unconditional}, we depict the results of unconditional sampling with FID metric on LSUN Bedroom and ImageNet datasets. In Table \ref{tab:class_conditional}, we show the results of class-conditional sampling with FID metric and different guidance scales. In Table \ref{tab:text_conditional}, we show the results on text-conditional sampling with human preference score $\uparrow$ with the human preference model HPD v2 \cite{wu2023human}.

\definecolor{lightgray}{gray}{0.5}
\setlength{\arrayrulewidth}{0.5pt}
\begin{table*}[htbp]
\small
\setlength{\tabcolsep}{10pt}
\caption{Sample quality of unconditional sampling measured by FID $\downarrow$ on LSUN Bedroom and ImageNet datasets, varying the number of function evaluations (NFE).} 
\label{tab:unconditional}
\centering
\begin{tabular}{clcccc}
	\toprule
        Dataset & Sampling Method $\backslash$ NFE  & 5 & 6 & 7 & 8   \\  
        \midrule
	\multirow{6}{*}{LSUN Bedroom}   & DDIM (Base)  & 51.482  & 32.470  & 22.533  & 17.775           \\
	  & DDIM (+Ours)  & 36.288  & 18.738  & 13.352  & 12.744                  \\
        \arrayrulecolor{lightgray}\cmidrule(lr){2-6}
	  & DPM-Solver (Base)  & 24.607  & 17.191  & 13.722  & 11.766            \\
	  & DPM-Solver (+Ours)  & 16.270  & 12.947  & 11.776  & 11.257            \\
        \arrayrulecolor{lightgray}\cmidrule(lr){2-6}
        \arrayrulecolor{black}
	  & DPM-Solver++ (Base)  & 24.378  & 16.705  & 13.414  & 11.635          \\
	  & DPM-Solver++ (+Ours)  & 16.594  & 13.295  & 12.416  & 11.218          \\   
        \midrule
	\multirow{6}{*}{ImageNet}   & DDIM (Base)  & 63.653  & 48.191  & 40.295  & 35.047           \\
	  & DDIM (+Ours)  & 49.379  & 32.729  & 26.147  & 23.084                 \\
        \arrayrulecolor{lightgray}\cmidrule(lr){2-6}
	  & DPM-Solver (Base)  & 35.673  & 28.797  & 24.729  & 22.400            \\
	  & DPM-Solver (+Ours)  & 28.353  & 23.653  & 21.261  & 19.974            \\
        \arrayrulecolor{lightgray}\cmidrule(lr){2-6}
        \arrayrulecolor{black} 
	  & DPM-Solver++ (Base)  & 35.118  & 28.342  & 24.298  & 22.003          \\
	  & DPM-Solver++ (+Ours)  & 28.602  & 24.022  & 21.607  & 20.486          \\
        \bottomrule
\end{tabular}
\end{table*}

\definecolor{lightgray}{gray}{0.5}
\setlength{\arrayrulewidth}{0.5pt} 
\begin{table*}[htbp]
\small
\setlength{\tabcolsep}{10pt}
\caption{Sample quality of class-conditional sampling measured by FID $\downarrow$ on ImageNet 256×256 \cite{dhariwal2021diffusion}, varying the number of function evaluations (NFE) and guidance scale.} 
\label{tab:class_conditional}
\centering
\begin{tabular}{clcccc}
	\toprule
        Guidance Scale & Sampling Method $\backslash$ NFE  & 5 & 6 & 7 & 8   \\  
        \midrule
	\multirow{6}{*}{2.0}   & DDIM (Base)  & 32.599  & 22.894  & 18.043  & 15.552           \\
	  & DDIM (+Ours)  & 26.018  & 17.316  & 13.798  & 12.460                  \\
        \arrayrulecolor{lightgray}\cmidrule(lr){2-6}
	  & DPM-Solver (Base)  & 16.549  & 12.901  & 10.907  & 9.807            \\
	  & DPM-Solver (+Ours)  & 14.046  & 11.210  & 9.902   & 9.240            \\
        \arrayrulecolor{lightgray}\cmidrule(lr){2-6}
        \arrayrulecolor{black} 
	  & DPM-Solver++ (Base)  & 16.753  & 12.962  & 10.960  & 9.799          \\
	  & DPM-Solver++ (+Ours)  & 14.675  & 11.701  & 10.296  & 9.582          \\   
        \midrule
	\multirow{6}{*}{4.0}   & DDIM (Base)  & 31.258  & 21.389  & 16.751  & 14.202           \\
	  & DDIM (+Ours)  & 29.265  & 18.645  & 14.401  & 12.554                  \\
        \arrayrulecolor{lightgray}\cmidrule(lr){2-6}
	  & DPM-Solver (Base)  & 16.909  & 12.548  & 10.685  & 9.707            \\
	  & DPM-Solver (+Ours)  & 15.759  & 11.790  & 10.163  & 9.380            \\
        \arrayrulecolor{lightgray}\cmidrule(lr){2-6}
        \arrayrulecolor{black} 
	  & DPM-Solver++ (Base)  & 17.381  & 12.832  & 10.842  & 9.791          \\
	  & DPM-Solver++ (+Ours)  & 16.695  & 12.350  & 10.546  & 9.605          \\
        \bottomrule
\end{tabular}
\end{table*}

\definecolor{lightgray}{gray}{0.5}
\setlength{\arrayrulewidth}{0.5pt} 
\begin{table*}[htbp]
\small
\setlength{\tabcolsep}{16pt}
\caption{Sample quality of text-conditional sampling measured by human preference score $\uparrow$ (human preference model HPD v2 \cite{wu2023human}) with captions from MS-COCO2014 validation dataset, varying the number of function evaluations (NFE).} 
\label{tab:text_conditional}
\centering
\begin{tabular}{lcccc}
	\toprule
        Sampling Method $\backslash$ NFE  & 5 & 10 & 15 & 20  \\  
        \arrayrulecolor{lightgray}\cmidrule(lr){1-5}
        \arrayrulecolor{black} 
	  DDIM (Base)  & 0.21492  & 0.25061  & 0.25831  & 0.26247           \\
	  DDIM (+Ours)  & 0.22828  & 0.25714  & 0.26168  & 0.26430                  \\
        \midrule
        Sampling Method $\backslash$ NFE  & 5 & 6 & 7 & 8   \\  
        \arrayrulecolor{lightgray}\cmidrule(lr){1-5}
        \arrayrulecolor{black} 
	  DPM-Solver (Base)  & 0.24097  & 0.25014  & 0.25461   & 0.25803            \\
	  DPM-Solver (+Ours)  & 0.24687  & 0.25412  & 0.25707   & 0.25904            \\
        \midrule
        Sampling Method $\backslash$ NFE  & 5 & 6 & 7 & 8   \\  
        \arrayrulecolor{lightgray}\cmidrule(lr){1-5}
        \arrayrulecolor{black} 
	  DPM-Solver++ (Base)  & 0.24146  & 0.25075  & 0.25521  & 0.25859          \\
	  DPM-Solver++ (+Ours)  & 0.24637  & 0.25330  & 0.25607  & 0.25944          \\   
        \bottomrule
\end{tabular}
\end{table*}

\section{Comparison with SDE-based sampler}    \label{sec:SDE}
\begin{figure}[htbp]
    \begin{center}
	\includegraphics[width=0.98\linewidth]{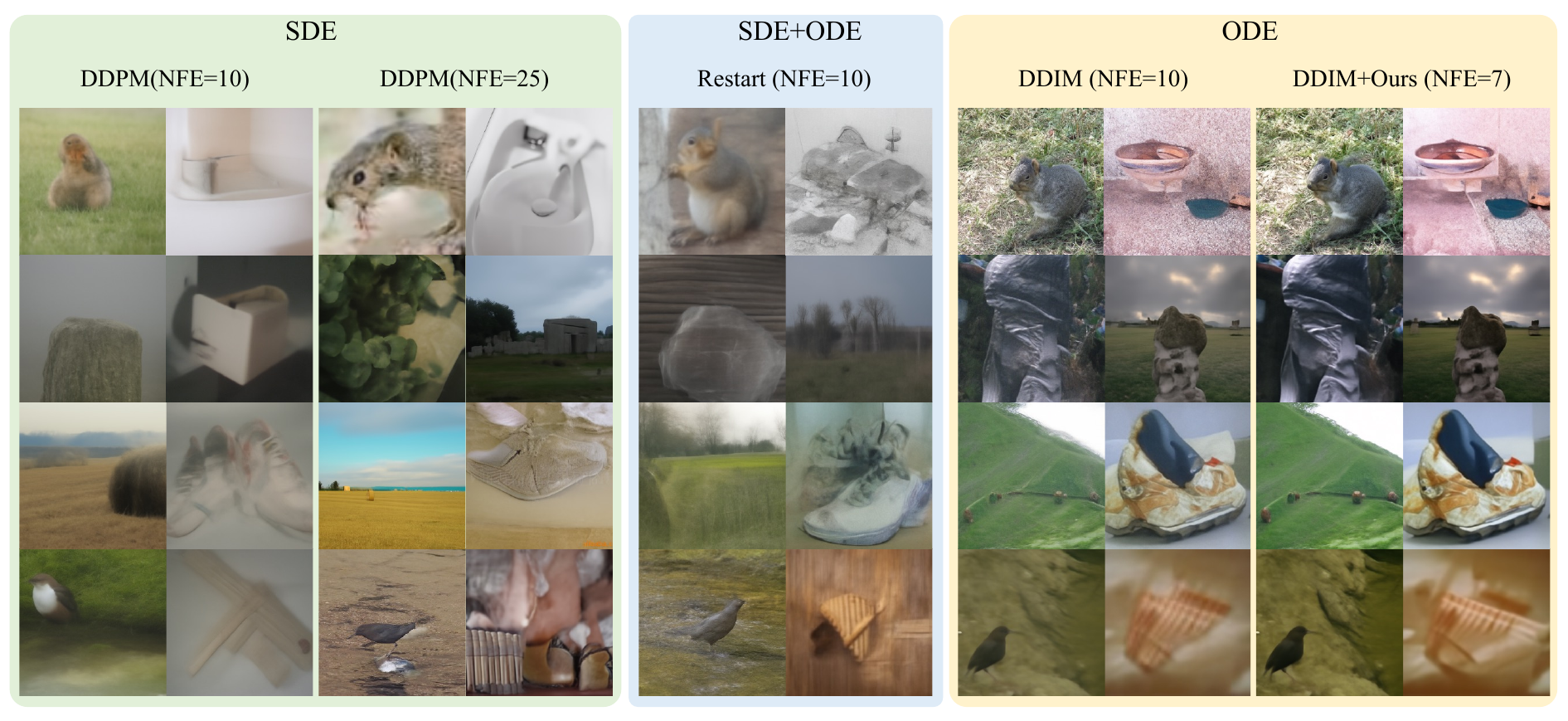}
    \end{center}
    \caption{Visual comparison with SDE-based samplers. All images are generated by sampling from a DPM trained on ImageNet 256×256 \cite{dhariwal2021diffusion}. Our method is superior to both SDE and ODE samplers in quality and speed, generating more plausible samples with more visual details and higher contrast.} 
    \label{fig:SDE}
\end{figure}

Large step size in stochastic differential equations (SDEs) violates the randomness of the Wiener process \cite{Kloeden1992numerical} and often causes non-convergence. Therefore, SDE-based sampler usually adopts hundreds of NFEs for inference. Certain methods \cite{guo2024gaussian,gonzalez2024seeds} propose to accelerate SDE solvers but still require hundreds of steps for inference. Restart \cite{xu2023restart} proposes to combine SDE and ODE via introducing stochasticity into the ODE process. These methods make promising attempts to accelerate SDE samplers, while still lag behind ODE solvers in speed. Besides, SDE-based sampler leads to stochastic generation, compared to the deterministic generation of ODEs. 
As shown in Figure \ref{fig:SDE}, we depict the visual comparison between these various samplers, including the SDE-based sampler DDPM \cite{ho2020denoising}, the SDE-ODE combined sampler Restart \cite{xu2023restart}, as well as the ODE-based sampler DDIM \cite{song2020denoising} and its enhanced version with our SynBoost. DDPM suffers from blurry results with small NFEs. Restart effectively elevates the speed of DDPM but still generates low-quality images with small NFEs. Besides, both DDPM and Restart lead to stochastic generation. In contrast, DDIM sampler performs better with finer details and structure. Further, with our SynBoost framework, the sampling quality and speed of DDIM are substantially boosted. 

\section{Exploring the upper bound of SynBoost}    \label{sec:bound}
\begin{figure}[htbp]
    \begin{center}
	\includegraphics[width=0.98\linewidth]{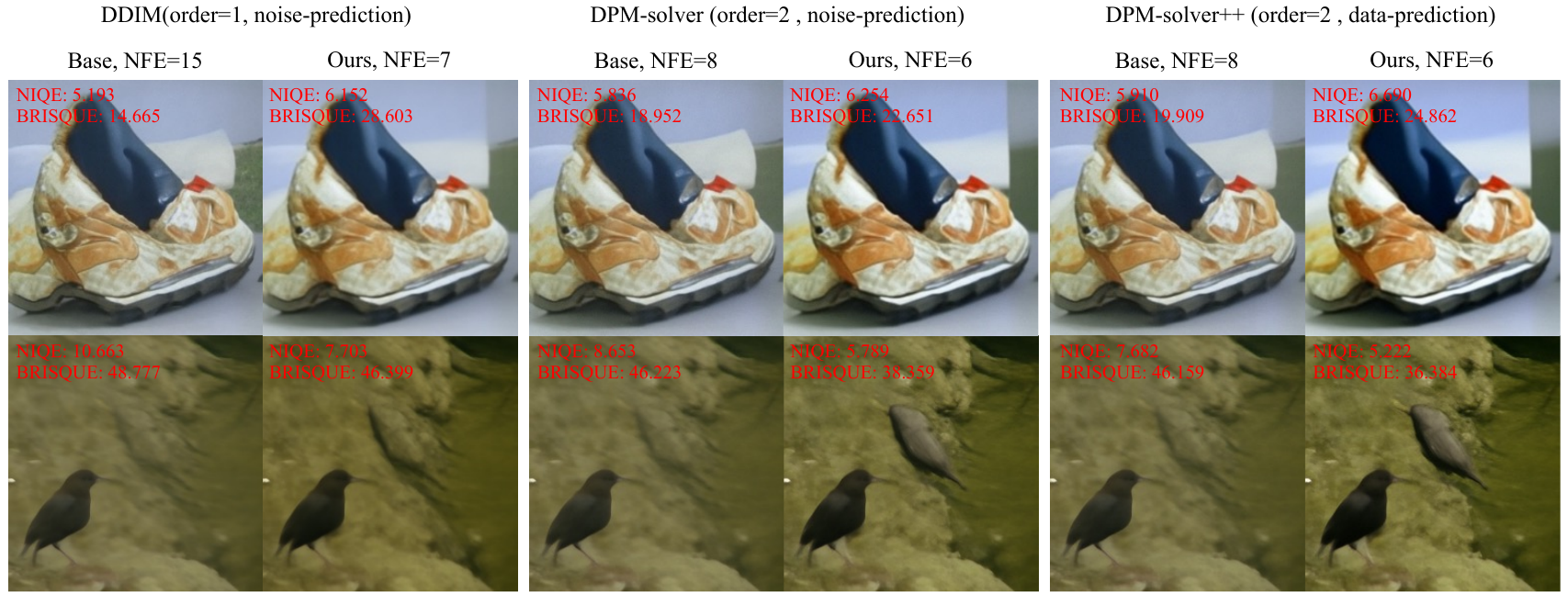}
    \end{center}
    \caption{The minimal steps required to generate visually clear images. All images are generated unconditionally. Our method can consistently lower the minimal steps for clear image generation. Best viewed in color. } 
    \label{fig:minimalstep}
\end{figure}

It is important and of practical value to explore the performance upper bound of existing samplers and our SynBoost for fast sampling. Concretely speaking, we desire to reveal the minimal sampling step required by existing samplers to generate visually clear and pleasing images. We adopt human preference as well as two well-known no-reference image quality assessment indicators: BRISQUE $\downarrow$ \cite{mittal2012no} and NIQE $\downarrow$ \cite{mittal2012making} to assess the visual results. As shown in Figure \ref{fig:minimalstep}, we depict the visual comparison under minimal sampling step of different samplers, and conclude two main conclusions. SynBoost achieves a larger minimal step reduction than increasing sampler order. For example, 2-order DPM-Solver reduces the minimal step from 15 to 8, compared to 1-order DDIM. While, our SynBoost enables DDIM to achieve minimal-step of 7.

Besides, SynBoost can also significantly reduce the minimal step requirement of high order samplers, like DPM-Solver and DPM-Solver++. For example, SynBoost further reduces the minimal sampling step of DPM-Solver from 8 to 6. This validates the generality and robustness of SynBoost.

\section{Sampling diversity}    \label{sec:diversity}
We also investigate the diversity of the images generated by SynBoost. In Table \ref{table:IS}, we compare the sampling diversity of SynBoost and base samplers with the inception score (IS) metric on ImageNet dataset. SynBoost can consistently improve the sampling diversity on various NFEs.

\begin{table*}[htbp]
\setlength{\tabcolsep}{6pt}
\caption{Comparisons of sampling diversity. We compute the IS score on ImageNet dataset, where SynBoost consistently improves the sampling diversity.}
\label{table:IS}
\centering
\begin{tabular}{c|cccc|cccc|cccc}
	\hline 
	\multirow{2}{*}{IS}     & \multicolumn{4}{|c}{DDIM}    & \multicolumn{4}{|c}{DPM-Solver}   & \multicolumn{4}{|c}{DPM-Solver++}   \\      \cline{2-13} 
	            & 5 & 6 & 7 & 8         & 5 & 6 & 7 & 8          & 5 & 6 & 7 & 8      \\ \hline
	Base        & 29.9 & 40.3 & 46.8 & 52.6 & 54.5 & 61.8 & 67.5 & 72.5 & 54.5 & 62.1 & 68.7 & 71.3  \\
	  Ours  & 38.7 & 54.4 & 64.2 & 69.9 & 61.6 & 70.0 & 74.3 & 75.4 & 58.6 & 67.1 & 71.4 & 72.7   \\ \hline
\end{tabular}
\end{table*}

\section{More Explanations of Equation 9 in the Manuscript} \label{sec:explanations}
Specifically, we detail: 1) its strict mathematical derivation from the exact DDIM sampling step; 2) the theoretical expansion of the direct substitution baseline; and 3) empirical evidence demonstrating the severe numerical instability (e.g., generation failure) inherently caused by direct substitution.

\begin{itemize}
    \item \textbf{Why mixing outperforms direct substitution.} 
    While $\epsilon_{\theta}(x_{\tau},\tau)$ provides a more accurate estimation of the score function for timestep $\tau$, completely replacing the current update direction $D_{t_i}$ with it would introduce a severe temporal mismatch. The network would ignore the local gradient at $t_i$. Our mixing formula (Eq. 9) elegantly avoids this by preserving the current step's data estimation while replacing only the noise estimation. 
    
    Specifically, the exact sampling formulation of DDIM is:
    \begin{equation}
        \small
        \label{eq:resp_ddim_base}
        x_{t_i}^{base}= \alpha_{t_{i}} x_{\theta}(x_{t_{i-1}}, t_{i-1}) + \sigma_{t_{i}} \epsilon_{\theta}(x_{t_{i-1}}, t_{i-1}).
    \end{equation}
    When expressed in the unified ODE form, the base update direction is:
    \begin{equation}
        \small
        \label{eq:resp_d_base}
        D_{t_{i}}^{base} = \epsilon_{\theta}(x_{t_{i-1}}, t_{i-1}).
    \end{equation}
    If we substitute the noise estimation part with the more accurate prior $\epsilon_{\theta}(x_{\tau}, \tau)$, but \textbf{strictly retain the current local data prediction} $x_{\theta}(x_{t_{i-1}}, t_{i-1})$, the modified sampling step becomes:
    \begin{equation}
        \small
        \label{eq:resp_x_ours}
        x_{t_i}^{ours}= \alpha_{t_{i}} x_{\theta}(x_{t_{i-1}}, t_{i-1}) + \sigma_{t_{i}} \epsilon_{\theta}(x_{\tau}, \tau).
    \end{equation}
    By algebraically rewriting this modified step back into the unified ODE solver format (Eq. 6), the corresponding update direction $D_{t_{i}}^{ours}$ naturally emerges as:
    \begin{equation}
        \small
        \label{eq:resp_d_ours}
        D_{t_{i}}^{ours} = \left(1 + \frac{1}{e^{h_i}-1}\right) \epsilon_{\theta}(x_{t_{i-1}}, t_{i-1}) - \frac{1}{e^{h_i}-1} \epsilon_{\theta}(x_{\tau}, \tau).
    \end{equation}
    This derivation proves that our mixing strategy is mathematically equivalent to leveraging the accurate noise prior while strictly grounding the transition in the current step's data prediction. The mixing coefficient $c = 1 / (e^{h_i}-1)$ is not arbitrary, but analytically derived from this process.

    \item \textbf{Mathematical flaw of direct substitution.}
    To provide deeper mathematical insight into why direct substitution fails, we can eliminate the step size variable $h_i$ and fully expand the update rule. In ODE-based methods, $h_i$ is defined as $h_i = \lambda_{t_i} - \lambda_{t_{i-1}}$, where $\lambda_t = \log\left(\frac{\alpha_t}{\sigma_t}\right)$. Consequently, $e^{h_i}$ can be expanded as:
    \begin{equation}
        \small
        e^{h_i} = \exp\left(\log\frac{\alpha_{t_i}}{\sigma_{t_i}} - \log\frac{\alpha_{t_{i-1}}}{\sigma_{t_{i-1}}}\right) = \frac{\alpha_{t_i} \sigma_{t_{i-1}}}{\sigma_{t_i} \alpha_{t_{i-1}}}.
    \end{equation}
    Substituting this into the step size coefficient $\sigma_{t_i}(e^{h_i}-1)$ yields:
    \begin{equation}
        \small
        \sigma_{t_i}(e^{h_i}-1) = \sigma_{t_i}\left(\frac{\alpha_{t_i} \sigma_{t_{i-1}}}{\sigma_{t_i} \alpha_{t_{i-1}}} - 1\right) = \frac{\alpha_{t_i} \sigma_{t_{i-1}}}{\alpha_{t_{i-1}}} - \sigma_{t_i}.
    \end{equation}
    By applying this coefficient, the direct substitution update rule becomes:
    \begin{equation}
        \small
        x_{t_i}^{direct} = \frac{\alpha_{t_i}}{\alpha_{t_{i-1}}} x_{t_{i-1}} - \left( \frac{\alpha_{t_i} \sigma_{t_{i-1}}}{\alpha_{t_{i-1}}} - \sigma_{t_i} \right) \epsilon_{\theta}(x_\tau, \tau).
    \end{equation}
    To compare this with the standard DDIM formulation ($x_{t_i} = \alpha_{t_i}x_\theta + \sigma_{t_i}\epsilon_\theta$), we fully expand $x_{t_{i-1}}$ as $x_{t_{i-1}} = \alpha_{t_{i-1}}x_\theta(x_{t_{i-1}}, t_{i-1}) + \sigma_{t_{i-1}}\epsilon_\theta(x_{t_{i-1}}, t_{i-1})$. Plugging this into the equation yields:
    \begin{equation}
    \small
        \label{supeq:3}
        \begin{aligned}
        x_{t_i}^{direct} = &\alpha_{t_i} x_{\theta}(x_{t_{i-1}}, t_{i-1}) + \frac{\alpha_{t_i} \sigma_{t_{i-1}}}{\alpha_{t_{i-1}}} \epsilon_{\theta}(x_{t_{i-1}}, t_{i-1}) \\ 
        &- \left( \frac{\alpha_{t_i} \sigma_{t_{i-1}}}{\alpha_{t_{i-1}}} - \sigma_{t_i} \right) \epsilon_{\theta}(x_\tau, \tau).
        \end{aligned}
    \end{equation} 

    This fully expanded equation mathematically exposes the critical flaw of direct substitution. Instead of preserving the elegant variance schedule of diffusion models (as shown in our clean $x_{t_i}^{ours}$ formulation), direct substitution results in a disjointed and mathematically unprincipled mixture of local noise $\epsilon_{\theta}(t_{i-1})$ and prior noise $\epsilon_{\theta}(\tau)$. 

    \item \textbf{Empirical evidence: numerical explosion.}
    Furthermore, when we implement Equation 9 via direct substitution, the generated images appear completely grey. This empirical failure perfectly corroborates our mathematical derivation and serves as strong evidence against direct substitution. 
    
    The root cause of this failure is severe numerical instability. Based on the expanded formula above, we define the coefficient for the noise term as $C = \frac{\alpha_{t_i} \sigma_{t_{i-1}}}{\alpha_{t_{i-1}}}$. In the early sampling stages (i.e., when $t$ is large, approaching $T=1000$), $\alpha_{t_{i-1}}$ is an extremely small value (e.g., $4 \times 10^{-5}$ or less). With $\alpha_{t_{i-1}}$ in the denominator, the coefficient $C$ explodes, approaching infinity. This causes the values within the tensor to overflow, producing NaNs or astronomically large numbers. During image saving, these anomalous values are forcibly clipped or normalized, inevitably resulting in the entirely grey images. This proves that direct substitution fundamentally destroys the numerical stability of the ODE solver.
\end{itemize}

\section{More Visual Results}    \label{sec:Visual}
In this part, we show more visual results with various samplers (DDIM, DPM-Solver, and DPM-Solver++), sampling types (unconditional, class-conditional, and text-conditional sampling), sampling spaces (pixel and latent space), NFEs (5, 6, 7, and 8), and guidance scales (0.0 and 2.0).

\begin{figure}[htbp]
    \begin{center}
	\includegraphics[width=0.98\linewidth]{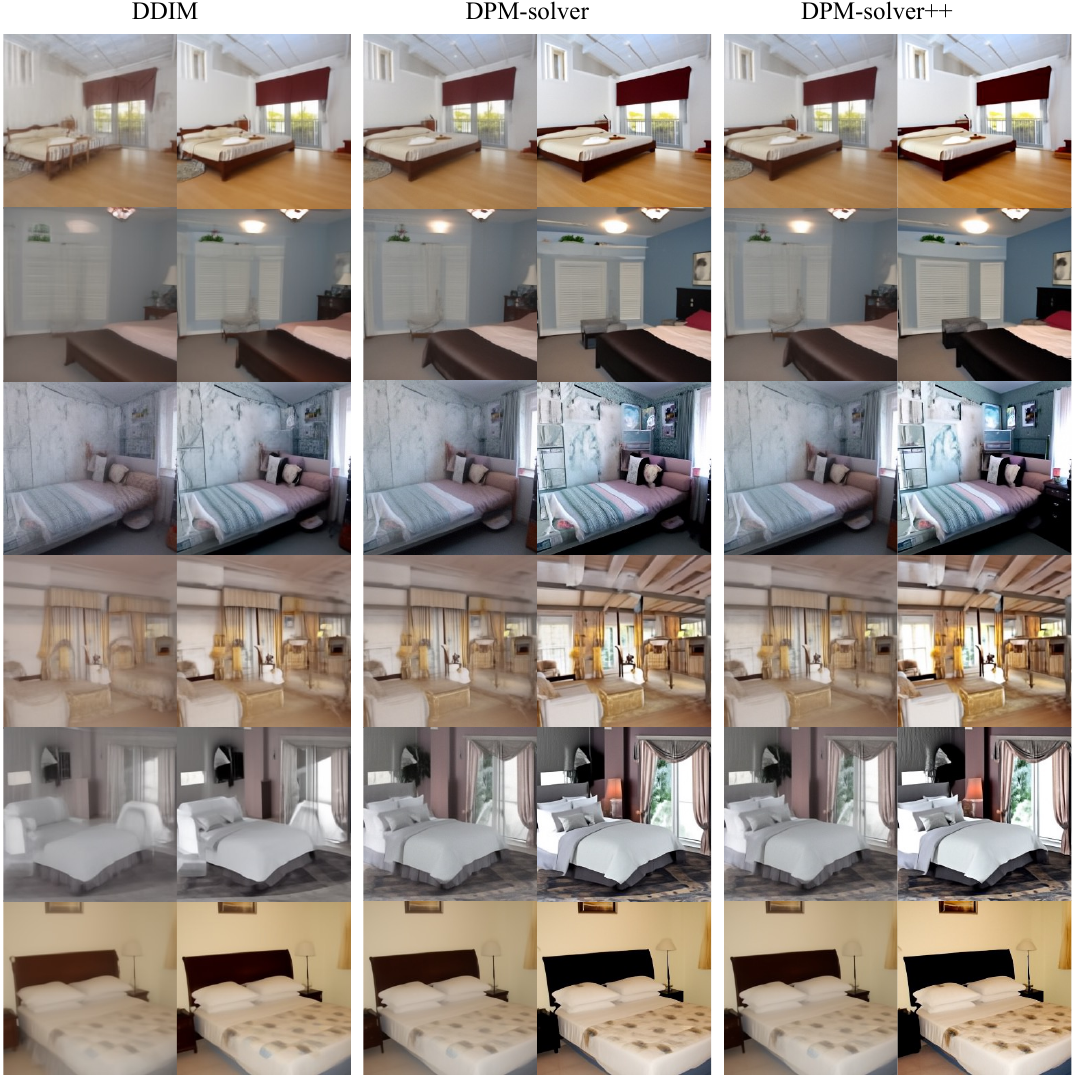}
    \end{center}
    \caption{Visual results of unconditional sampling on LSUN Bedroom dataset.} 
    \label{fig:sup_lsun}
\end{figure}

\begin{figure}[htbp]
    \begin{center}
	\includegraphics[width=0.98\linewidth]{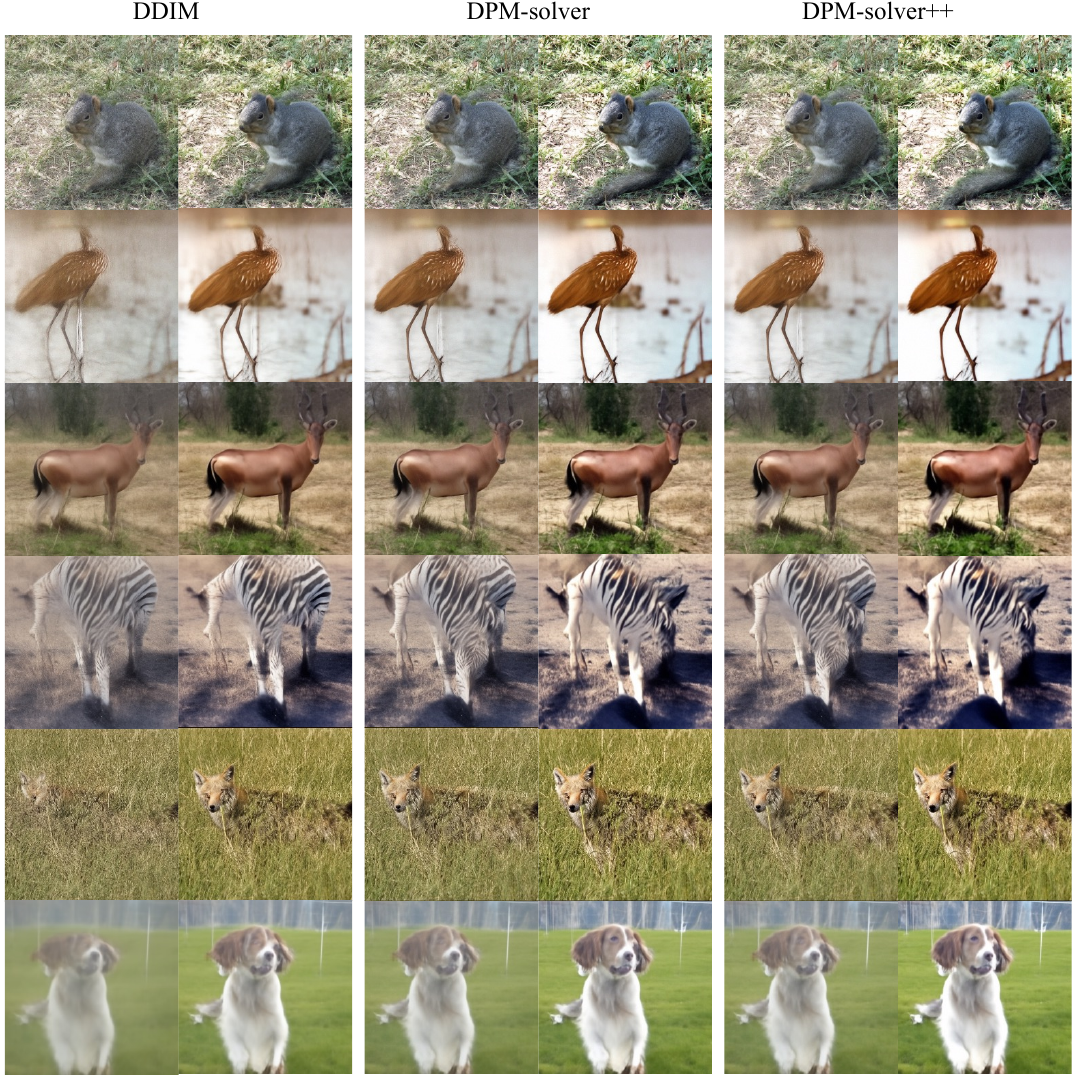}
    \end{center}
    \caption{Visual results of unconditional sampling on ImageNet dataset.} 
    \label{fig:sup_pixel_s0}
\end{figure}

\begin{figure}[t]
    \begin{center}
	\includegraphics[width=0.98\linewidth]{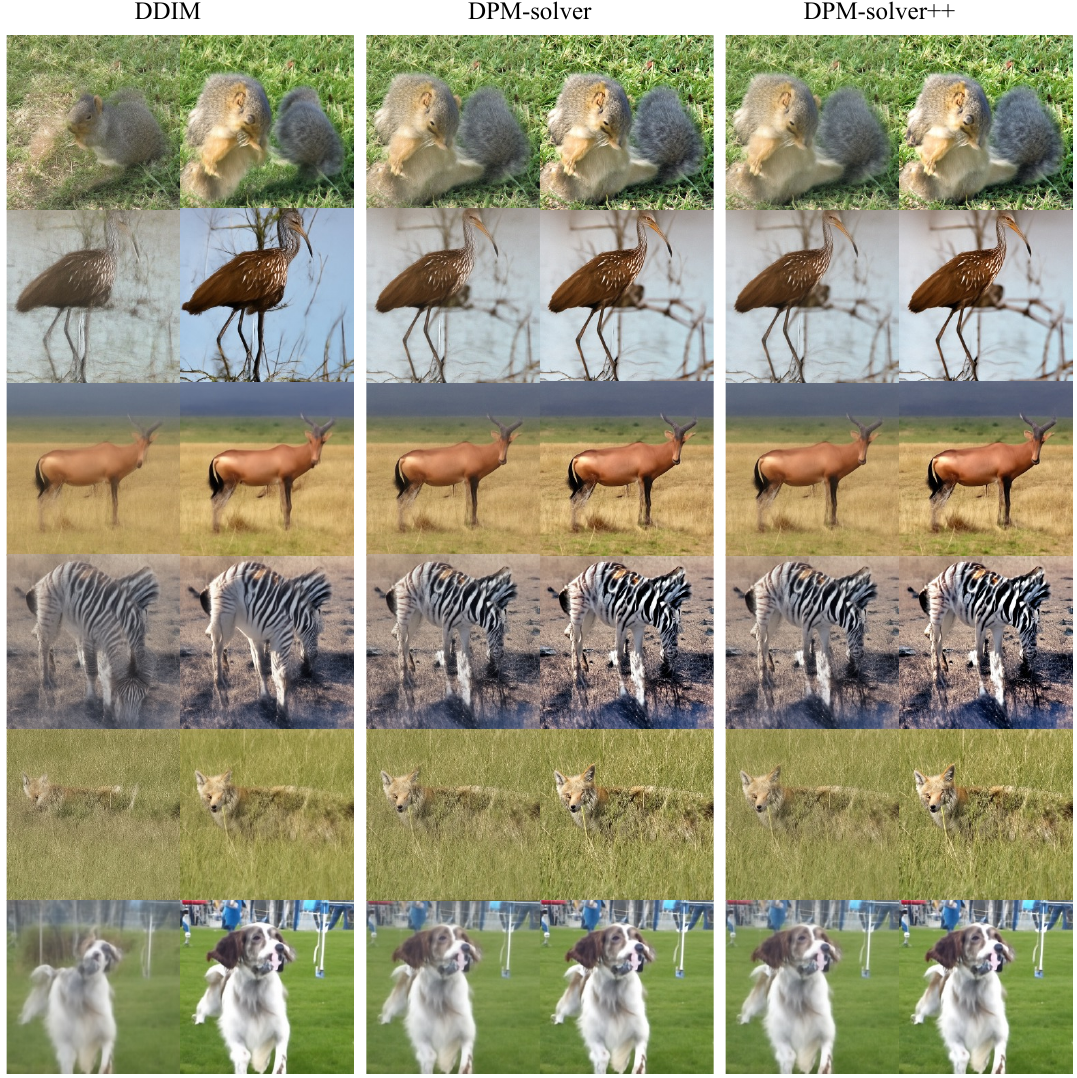}
    \end{center}
    \caption{Visual results of class-conditional sampling on ImageNet dataset with guidance scale 2.0.}
    \label{fig:sup_pixel_s2}
\end{figure}

\end{document}